\documentclass[lettersize,journal]{IEEEtran}
\usepackage{amsmath,amsfonts}
\usepackage{algorithmic}
\usepackage{algorithm}
\usepackage{array}
\usepackage[caption=false,font=normalsize,labelfont=sf,textfont=sf]{subfig}
\usepackage{textcomp}
\usepackage{stfloats}
\usepackage{url}
\usepackage{verbatim}
\usepackage{graphicx}
\usepackage{cite}
\usepackage{siunitx}
\usepackage{amsmath,amssymb,amsfonts}
\usepackage{algorithmic}
\usepackage{wrapfig}
\usepackage{booktabs}
\usepackage{threeparttable}
\usepackage{multirow}
\usepackage{comment}

\begin{document}

\title{BenthicDINO: Physics-Informed Self-Distillation for View-Invariant Side-Scan Sonar Representations}

\author{Taqi Hamoda,
        Hayat Rajani,~\IEEEmembership{Member,~IEEE,}
        and Nuno Gracias,~\IEEEmembership{Member,~IEEE}
        % <-this % stops a space

\thanks{All authors are with the Computer Vision and Robotics Research Institute (ViCOROB) of the University of Girona, Spain.}% <-this % stops a space

\thanks{Corresponding author: Hayat Rajani (hayat.rajani@udg.edu).}}



\maketitle

\begin{abstract}
Automated interpretation of side-scan sonar (SSS) imagery is traditionally hindered by view-dependence, as recorded intensities inextricably conflate intrinsic seabed reflectivity with transient acquisition geometries and acoustic artifacts. 
To address this, we introduce BenthicDINO, a physics-informed self-supervised learning framework that decouples acquisition dynamics from seabed representations. Utilizing a ConvNeXt-v2 backbone, physics-guided augmentations, and an HSIC penalty, BenthicDINO explicitly enforces feature invariance to physical viewing parameters like slant range and incidence angle.
Evaluations demonstrate exceptional data efficiency—achieving near-peak performance in benthic habitat mapping with only 10\% of available labels—and strong cross-domain generalization, outperforming fully supervised baselines on the AI4Shipwrecks benchmark using frozen representations.
\end{abstract}


\section{Introduction}

\IEEEPARstart{U}{nderwater} optical perception is fundamentally constrained by the physics of the marine environment. Electromagnetic radiation attenuates rapidly in water due to light absorption and scattering, limiting high-resolution optical sensors to ranges below 10 \si{m} \cite{burguera2016, sss_book}. In contrast, acoustic energy propagates with low attenuation, allowing sound waves to travel hundreds of meters in turbid and low-light conditions \cite{burguera2016, sss_handbook}; as such, side-scan sonar (SSS) has become the de-facto primary modality for large-scale seafloor mapping, long-range target detection, and autonomous navigation \cite{burguera2016, sss_book, sss_handbook}.

Unlike optical sensors, SSS image formation is directly governed by acoustic acquisition dynamics. Laterally mounted transducers on a moving platform emit fan-shaped pulses perpendicular to the trajectory. Returning echoes are recorded over time and converted to range assuming a constant sound speed of \SI{1500}{\meter\per\second}, yielding a continuous 2D backscatter intensity map \cite{burguera2016, sss_book, sss_handbook}. Traditionally, this process is approximated by the Lambertian model:

\begin{equation}
I(x,y) = \rho(x,y) \cdot \cos(\theta(x,y)) \cdot L(x,y)
\end{equation}

where the recorded intensity $I(x,y)$ is a function of the intrinsic seabed reflectivity $\rho(x, y)$, the local incidence angle $\theta(x, y)$, and the acoustic propagation loss $L(x, y)$ \cite{coiras2007}.

\begin{figure}[!t]
    \centering
    \includegraphics[width=\columnwidth]{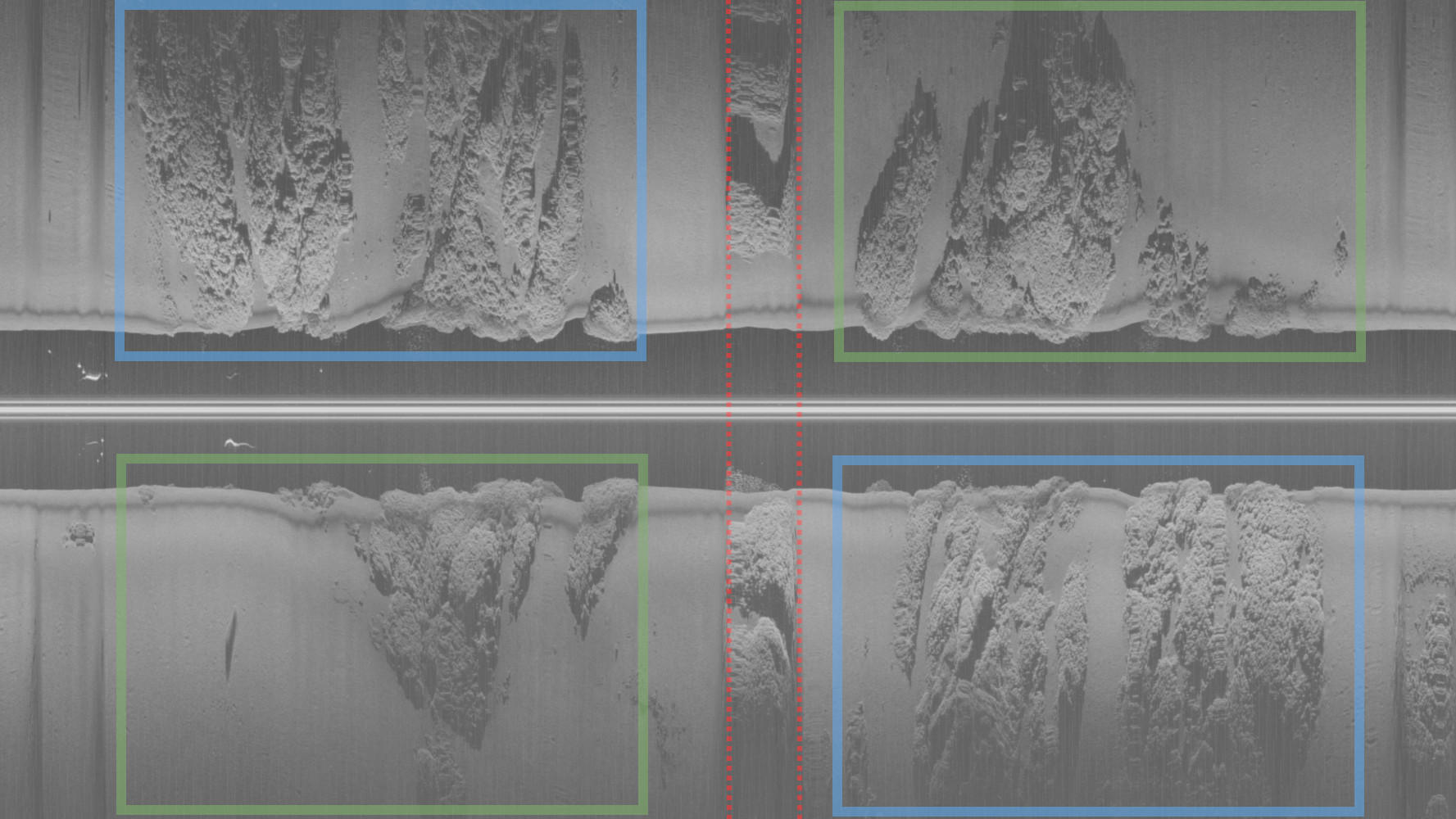}
    \caption{SSS waterfall image showing a heading reversal midway through the survey. Blue and green boxes highlight the same physical seabed features imaged from opposite views. Vertical red dashed lines indicate $90^{\circ}$ clockwise turns executed by the vehicle.}
    \label{fig:view_dependence}
\end{figure}

Because a significant portion of the recorded intensity encodes the acquisition geometry rather than the seabed itself, features extracted from raw imagery inextricably conflate intrinsic reflectivity ($\rho$) with transient viewing conditions. Consequently, SSS imagery is highly view-dependent; as demonstrated in Fig.~\ref{fig:view_dependence}, the exact same seafloor structure can produce drastically different feature scales and shadow geometries when surveyed from alternative directions or ranges.

These imaging artifacts severely bottleneck automated perception. Viewpoint dependence and non-uniform intensity result in poor repeatability for traditional handcrafted key-point extractors (e.g. SIFT \cite{sift}, SURF \cite{surf}), causing high matching errors and rendering real-time analytical interpretation infeasible \cite{can2025geo_match, katrusha2025, fu2025}. The most direct consequence is unreliable cross-view correspondence: when the same location is imaged from two different headings, the views are difficult to associate, weakening loop closure, mosaicking, and change detection over repeated surveys. A view-invariant representation \textemdash one that responds to the structural seabed and not the sensor geometry \textemdash is therefore a prerequisite for robust large-scale SSS perception.

To overcome these limitations, recent research has turned towards deep self-supervised learning (SSL) \cite{alan2022, sheffield2023advances, can2025_physdnet, minejepa2026}. By leveraging large-scale unlabelled sonar data, SSL frameworks extract invariant features without relying on scarce manual annotations \cite{caron2021, asifullah2025}. Modern self-distillation methods like DINOv3 \cite{oriane2025} promote generic invariance through multi-crop augmentation. However, these standard augmentations are designed for optical camera sensors and fail to model the physical causes of SSS view dependence; consequently, the learned features still absorb geometric information. No existing SSL approach for SSS explicitly forces the learned representations to be statistically independent of the acquisition parameters.

We address this gap with a physics-guided self-supervised framework built upon the DINOv3 framework. We replace the default Vision Transformer (ViT) with a ConvNeXt-v2 Tiny \cite{woo2023} backbone, leveraging its convolutional inductive biases and global response normalization to stabilize masked latent learning on small SSS datasets. The framework operates on two complementary fronts. First, physics-guided augmentations reproduce dominant SSS nuisance factors during training: bounded additive Gaussian perturbations act as a de-noising regularizer against speckle, linear synthetic Time-Varying Gain (TVG) profiles enforce invariance to uncompensated attenuation, and radiometric jitter simulates gain miscalibration. Second, a Hilbert-Schmidt Independence Criterion (HSIC) \cite{hsic} penalty explicitly minimizes the statistical dependence between the learned dense patch features and the measured viewing parameters, effectively scrubbing residual geometry information that augmentations alone cannot suppress. Finally, a novel hierarchical MLP fusion strategy aggregates features across all four network stages, preserving high-frequency sediment details alongside deep semantic abstractions.

The main contributions of this work are as follows:
\begin{itemize}
\item We adapt the DINOv3 framework to acoustic data by integrating a ConvNeXt-v2 Tiny backbone and proposing a dense, multi-scale hierarchical feature fusion that retains fine-grained seabed textures.
\item We formulate view-invariance in SSS as a statistical independence problem, introducing an HSIC-based objective that explicitly decouples learned dense representations from acquisition geometries (slant range and incidence angle).
\item We design a set of physics-guided augmentations derived from the SSS acquisition model, safely injecting speckle, linear range-dependent attenuation, and radiometric variation to enforce physical invariances without manual annotation.
\item We demonstrate through extensive representation-quality analyses that the resulting embeddings are view-invariant, significantly improving cross-view correspondence reliability compared to existing self-supervised baselines, while transferring effectively to downstream perception tasks.
\end{itemize}
 
\section{Related Work}

\subsection{Representation Learning for Sonar}
Self-supervised learning (SSL) has seen increasing adoption for sonar perception, driven by the high cost of expert annotations and the poor transferability of models pretrained on natural optical images \cite{valdenegro2021pre}. Early investigations explored fundamental pretext tasks \textemdash such as rotation prediction, de-noising auto-encoders, and jigsaw solving \textemdash demonstrating that SSL significantly narrows the performance gap with supervised methods in low-label regimes for forward-looking sonar (FLS) \cite{alan2022}. For synthetic-aperture sonar (SAS), subsequent works adapted momentum-contrast pretraining \cite{sheffield2023advances} and surveyed both contrastive and generative SSL paradigms for processing and recognition \cite{sheffield2023advances}.

Within the SSS domain, recent research has evaluated modern architectures, comparing ViTs against convolutional networks (including ConvNeXt variants) for classification, highlighting SSL as a critical next step for acoustic data \cite{sheffield2024vit}. Concurrently, joint-embedding predictive architectures have been applied for in-domain mine-like object classification \cite{minejepa2026}. More advanced self-distillation frameworks, such as DINOv3, have proven highly effective at generating robust dense features \cite{oriane2025}, inspiring related FLS applications that leverage self-supervised feature-space transformations to suppress speckle and enhance target regions \cite{zhang2025fls}. 

Despite this rapid progress, existing sonar SSL pipelines inherit augmentations and pretext tasks designed for natural imagery. None explicitly enforce statistical invariance to the physical acquisition geometry or acoustic degradation models. This fundamental gap motivates our approach, which uniquely embeds an explicit independence constraint and physics-guided augmentations directly within a dense self-distillation framework.

\subsection{Physics-Informed Learning for Sonar}
The scarcity of paired, high-quality real sonar data has driven significant interest in physics-informed learning, largely focused on realistic image synthesis and closing the domain gap between simulated and real imagery. Explicit acoustic simulators, such as S3Simulator \cite{kamal2024s3sim} and ACOUSIM \cite{acousim2026}, model the acquisition process directly to generate benchmark datasets or measure statistical alignment without generative models. Another major line of work combines learned generators with physical priors to augment small datasets: Li \textit{et al.} explored zero-shot and few-shot SSS synthesis for target detection \cite{li2024side}, Peng \textit{et al.} trained multi-view generative adversarial networks \cite{peng2025multi}, Koo \textit{et al.} utilized CycleGANs for synthetic generation \cite{koo2024cyclegan}, and Ma \textit{et al.} embedded random-fusion strategies within diffusion frameworks for segmentation \cite{ma2025diffusion}.

A third group of methods grounds the physics directly in the acquisition geometry to normalize or transform backscatter intensity. For instance, Liu and Ye proposed a gray-scale correction method accounting for rugged seafloors \cite{liu2023gray}, Stewart \textit{et al.} mapped SAS intensity to seabed elevation via image-to-height translation \cite{stewart2023height}, and recent works have employed physics-based backscatter correction to reduce view-dependent variations prior to feature extraction \cite{can2025_physdnet}. 

While these approaches demonstrate that encoding the sensor model improves data realism and mitigates intensity inconsistencies, they universally apply physics during pre-processing or image synthesis. They do not embed physical constraints directly into the representation learning objective. Consequently, the resulting deep features are not mathematically guaranteed to be invariant to the acquisition geometry. Our work departs from this paradigm by enforcing geometric view-invariance directly in the latent space during pretraining, ensuring the network natively decouples intrinsic seabed structure from transient viewing conditions.

\section{Methodology}
We build on the DINOv3 self-distillation framework and adapt it to the acoustic characteristics of SSS. The framework has three parts: a base self-distillation model with a ConvNeXt-v2 Tiny encoder (Sections~\ref{sec:base} and \ref{sec:encoder}); a set of physics-guided augmentations that reproduce the dominant SSS nuisance factors during training (Section~\ref{sec:aug}); and an explicit statistical-independence penalty that decouples the learned dense features from the acquisition geometry (Section~\ref{sec:viewinv}). Furthermore, we introduce a set of training adaptations that make self-distillation stable under small-batch distributed training on SSS data that are described later with the experimental settings (Section~\ref{sec:training}).

\subsection{DINOv3 Self-Distillation Framework}
\label{sec:base}

The DINO family of methods employ a self-distillation paradigm where a student network learns to match the output distributions of a teacher network \cite{oriane2025, caron2021, oquab2024}. 
The teacher's weights are updated using an exponential moving average (EMA) of the student's weights, ensuring a stable target representation \cite{oriane2025, caron2021, oquab2024}.
The overall pretraining objective for DINOv3 is a composite loss function designed to capture both global semantic understanding and fine-grained local structures \cite{oriane2025}:

$$
\mathcal{L}_{Pre} = \mathcal{L}_{DINO} + \mathcal{L}_{iBOT} + 0.5 \cdot \mathcal{L}_{Gram} + 0.1 \cdot \mathcal{L}_{KoLeo}
$$

\subsubsection{Global Discriminative Loss}
The DINO objective enforces view-invariance at the global image level by aligning representations across different augmented views of the SSS input.
Specifically, the student network processes a comprehensive set of global and local crops, denoted as $V$, while the teacher network is restricted to processing only global crops $V_{\text{global}}$ \cite{oriane2025}.
The outputs from both networks are transformed into probability distributions over $K$ dimensions via a projection head followed by a softmax activation.
To effectively prevent representational collapse, Sinkhorn-Knopp centering is used for the teacher's distributions \cite{oriane2025}, replacing the original centering and sharpening mechanism used in DINOv1 \cite{caron2021, oquab2024}. The resulting loss minimizes the cross-entropy between the student distribution $P_s(x_v)$ and the centered teacher distribution $P_t(x_{v'})$.

$$
\mathcal{L}_{\text{DINO}} = -\sum_{v \in V} \sum_{v' \in V_{\text{global}}} P_t(x_{v'}) \log P_s(x_v)
$$

Minimizing this objective promotes the learning of global features that are robustly invariant to geometric distortions and sensor noise.

\subsubsection{Masked Latent Reconstruction}
To instill dense prediction capabilities and enhance the model's robustness to acoustic shadows, an iBOT masked image modeling loss is used to complement the global objective \cite{oquab2024}. This patch-level objective forces the student network to reconstruct the features of masked tokens using the teacher's representations of the corresponding unmasked patches as learning targets. Consistent with the global DINO objective, Sinkhorn-Knopp centering is also applied to these patch-level projections \cite{oriane2025}. For a given set of masked patch indices $\mathcal{M}$, the reconstruction loss is formulated as the cross-entropy between the teacher's patch distribution $P_{t_i}^{patch}$ and the student's prediction $P_{s_i}^{patch}$.

$$
\mathcal{L}_{iBOT} = - \sum_{i \in \mathcal{M}} P_{t_i}^{patch} \log P_{s_i}^{patch}
$$

\subsubsection{Feature Regularisation}
To prevent feature collapse and encourage a uniform distribution across the hypersphere, KoLeo regularization is applied on the dense embeddings of the student patches \cite{oquab2024, oriane2025}.

\begin{equation}
\mathcal{L}_{\text{KoLeo}} = -\frac{1}{n} \sum_{i=1}^{n} \ln \left( \min_{j \neq i} \| z_i - z_j \| \right)
\end{equation}

This term maximises the minimum distance between samples, which improves class separability in downstream benthic tasks. In $\mathcal{L}_{\text{Pre}}$ we use its distributed variant, which computes the same quantity across the distributed batch.

\subsubsection{Gram Anchoring}
Long training schedules and high resolutions often degrade dense feature maps, causing patch-level inconsistency and noisy representations that harm SSS matching and registration \cite{oriane2025}. DINOv3 introduces Gram anchoring to preserve local structure \cite{oriane2025}. The Gram matrix captures the pairwise correlations of the L2-normalised patch features. The student is regularised by matching its Gram matrix to that of a stable ``Gram teacher'', a checkpoint from earlier training.

\begin{equation}
\mathcal{L}_{\text{Gram}} = | X_S X_S^\top - X_G X_G^\top |_F^2
\end{equation}

Gram anchoring stabilises dense features, accelerates iBOT convergence, and improves performance on dense vision tasks such as feature matching and segmentation, while preserving fine image textures \cite{oriane2025}.

\begin{figure*}[t]
    \centering
    \includegraphics[width=\textwidth]{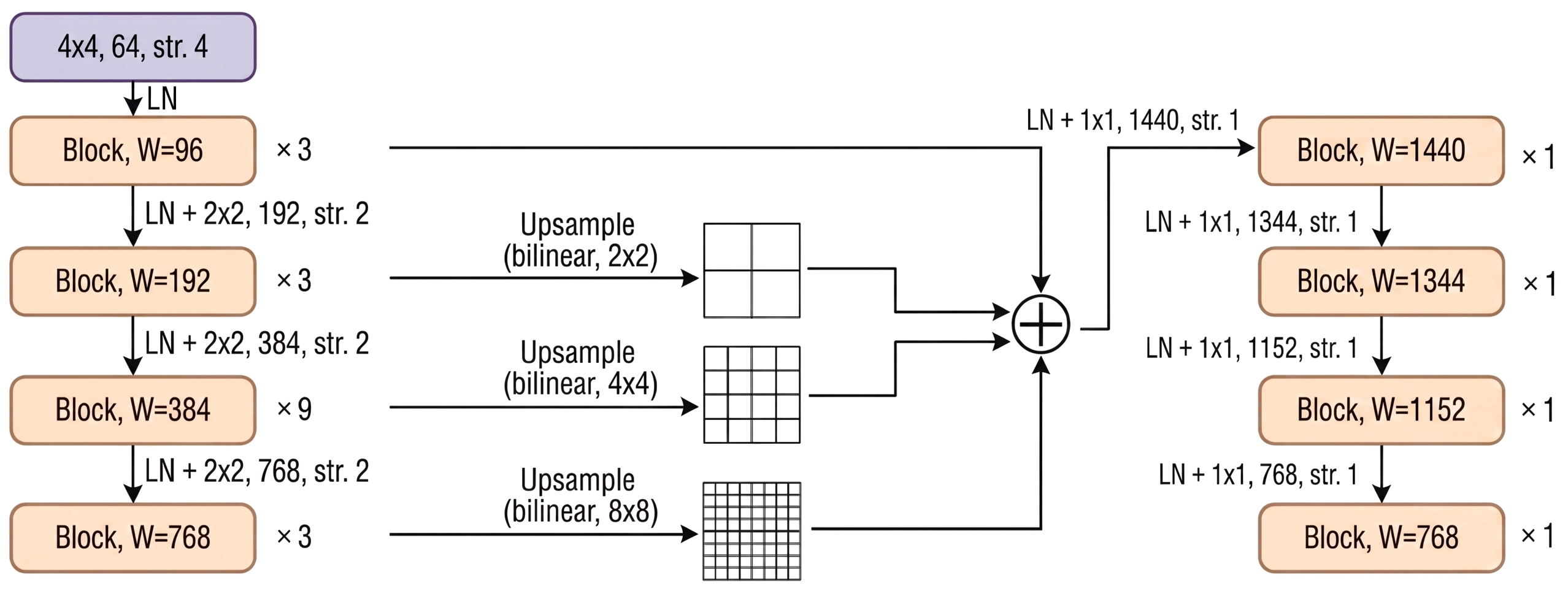}
    \caption{ConvNeXt-v2 adapted for DINOv3. Outputs from stages 2–4 are upsampled to match stage 1, concatenated, and compressed via a fusion MLP into a 768-D embedding.}
    \label{fig:arch}
\end{figure*}

\subsection{ConvNeXt-v2 Backbone}
\label{sec:encoder}

Standard implementations of the DINO framework typically rely heavily on ViTs as the backbone \cite{caron2021, oquab2024, oriane2025}.
However, ViTs lack the strong inductive biases inherent to CNNs, such as translation equivariance and local connectivity, and therefore demand massive amounts of training data to generalize effectively \cite{cnn_inductive, liu2022}.
Given that high-quality, diverse SSS data is notoriously scarce and our dataset is relatively small, we opted for a convolutional architecture to maximize learning efficiency.
Furthermore, selecting the \textit{tiny} variant was a deliberate regularization tactic; its constrained parameter space prevents the model from severely over-fitting to the limited acoustic dataset while remaining computationally viable for AUV deployment \cite{liu2022}.

The original ConvNeXt \cite{liu2022} modernized the standard ResNet block by incorporating ViT-inspired design choices, including large $7\times 7$ depthwise convolutions, inverted bottlenecks, and replacing Batch Normalization with Layer Normalization \cite{liu2022}.
While highly effective for supervised learning, applying mask-based self-supervised learning, such as the iBOT objective, directly to ConvNeXt induces severe representation collapse \cite{woo2023}.  
When trained on masked inputs, the dimension-expansion MLP layers produce dead or saturated feature maps, leading to redundant channel activations and degraded feature diversity \cite{woo2023}. 

To mitigate these issues, ConvNeXt-v2 enables mask-based training via sparse convolutions and Global Response Normalization (GRN) \cite{woo2023}.
Treating inputs as sparse point clouds, it applies submanifold sparse convolutions exclusively to visible patches.
Because this sparsity disrupts activation statistics, causing dead feature channels in MLP layers, ConvNeXt-v2 introduces GRN directly after the MLP.
By aggregating spatial features into a global $L_2$-norm vector and applying divisive normalization across channels \cite{woo2023}, GRN enforces inter-channel competition to revive dead channels and maintain feature diversity during masked reconstruction.

Furthermore, moving beyond previous multi-scale approaches that discarded early features, our framework extracts and aggregates hierarchical representations from \textit{all} four stages of the network as can be seen in Fig.~\ref{fig:arch}.
Let $F^{(s)}$ denote the feature map from stage $s \in \{1, 2, 3, 4\}$.  
Feature maps from stages 2, 3, and 4 are upsampled via bilinear interpolation to match the spatial resolution of stage 1; the upsampled feature maps are then concatenated along the channel dimension.
To effectively fuse these multi-scale representations without losing spatial context, we introduce a 3-layer Multi-Layer Perceptron (implemented via $1\times 1$ convolutions) that compresses the concatenated features into a unified 768-dimensional descriptor: 
$$
\mathcal{H}_{i,j} = \text{MLP}_{1\times1} \left( \bigoplus_{s=1}^{4} F^{(s)}_{i,j} \right) \in \mathbb{R}^{768}
$$
 This dense feature fusion effectively preserves the fine-grained, high-frequency sediment details captured in the initial stage, while seamlessly integrating the deep, robust semantic abstraction derived from the deeper stages. 

\begin{figure*}[t]
    \centering
    \includegraphics[width=\textwidth]{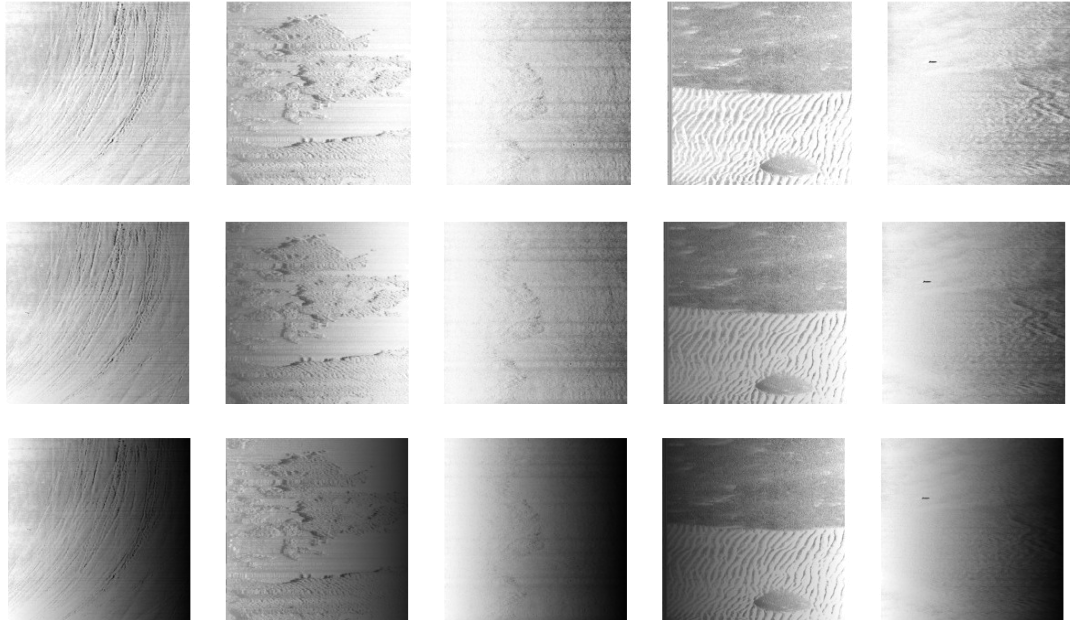}
    \caption{Effects of TVG Attenuation on five sample SSS patches, progressing from the original patches (top row) to moderate (middle row) and extreme (bottom row) attenuation levels.}
    \label{fig:tvg}
\end{figure*}

\subsection{Physics-Guided Augmentations}
\label{sec:aug}

DINOv3 induces its target invariances through the augmentations applied to each crop. The default photometric augmentations are designed for natural photographs and do not accurately represent the physical factors that affect SSS imagery. We therefore incorporate three physically motivated augmentations, each targeting a dominant SSS nuisance factor: speckle noise, range-dependent attenuation, and radiometric miscalibration. 

Each operation is applied independently to every crop. Consequently, the student and teacher process varying acoustic conditions for the same seabed patch, compelling the network to learn features invariant to these physical artifacts. Let $I \in \mathbb{R}^{H \times W}$ denote an input tile, with $x$ representing the along-track axis and $r$ representing the across-track (range) axis.

\subsubsection{Speckle Perturbation}
Coherent acoustic imaging produces speckle, a grainy multiplicative fluctuation that corrupts the backscatter signal \cite{alan2022}. To improve robustness to this degradation, we perturb the tile with additive Gaussian noise: 

\begin{equation}
\tilde{I}(x,r) = I(x,r) + \eta(x,r), \quad \eta(x,r) \sim \mathcal{N}(0, \sigma^2)
\end{equation}

By log-compressing raw intensities, the additive formulation becomes mathematically equivalent to the multiplicative method. This acts as a de-noising regularizer, forcing the network to prioritize robust structural representations over raw pixel values \cite{alan2022}.
Unlike usual assumptions of a zero-based uniform distribution, our implementation explicitly bounds the standard deviation to prevent catastrophic signal destruction. Specifically, $\sigma$ is sampled per tile from a uniform distribution $\sigma \sim \mathcal{U}(0.01, 0.05)$, and the augmentation is applied with a probability of $p=0.5$. 

\subsubsection{Synthetic Time-Varying Gain (TVG)}
Recorded acoustic intensity decays with range due to geometric spreading and absorption. While hardware TVG is normally applied to compensate for this loss, the compensation is often imperfect, leaving residual range-dependent trends that vary drastically between survey passes \cite{sss_handbook, burguera2016}. To enforce invariance to uncompensated transmission loss, we multiply the tile by a synthetic residual gain curve.

Our augmentation simulates this artifact by applying a linear decay gradient across the range of the log-compressed tile: 

\begin{equation}
\tilde{I}(x,r) = g(r) \cdot I(x,r), \quad g(r) = 1.0 - (1.0 - \beta) \frac{r}{r_{\max}}
\end{equation}

where $r_{\max}$ is the maximum range of the tile, and $\beta \sim \mathcal{U}(0.3, 0.7)$ represents the retention factor at the farthest edge (i.e., the far edge retains between $30\%$ and $70\%$ of its original intensity).
Applied with a probability of $p=0.5$, this linear decay simulates severe, uncompensated profile fading, driving the network to decouple structural features from spatial acoustic attenuation. This is as depicted in Fig.~\ref{fig:tvg}.

\subsubsection{Radiometric Jitter}
Operator adjustments and automatic gain control (AGC) consistently change the overall brightness and contrast of a survey, altering the dynamic range independently of the underlying seabed structure \cite{burguera2016}.
To make the representations invariant to these radiometric calibration errors, we apply a global brightness and contrast perturbation using standard jitter implementations:
\begin{equation}
\tilde{I}(x,r) = \kappa \big( I(x,r) - \bar{I} \big) + \bar{I} + \delta
\end{equation}
where $\bar{I}$ is the mean intensity of the log-compressed tile, $\kappa \sim \mathcal{U}(1-c, 1+c)$ is a contrast scaling factor with $c = 0.2$, and $\delta \sim \mathcal{U}(-b, b)$ is a brightness offset with $b = 0.1$.
By artificially perturbing the global intensity scale, the network is forced to isolate the intrinsic seabed reflectivity ($\rho$) rather than relying on transient intensity shifts \cite{can2025geo_match}.
The output is subsequently clipped to maintain valid, normalized intensity boundaries.

\subsection{Enforcing View-Invariance}
\label{sec:viewinv}

While the self-distillation framework of DINOv3 inherently promotes a degree of invariance through multi-crop augmentations, explicitly decoupling the learned representations from the physical viewing geometry can further isolate the intrinsic seafloor properties.
The physics-guided augmentations above do encourage invariance, but only implicitly. Residual geometric information can still leak into the features.
To remove it, we enforce statistical independence between the learned representations and the viewing parameters using the Hilbert-Schmidt Independence Criterion (HSIC) \cite{hsic}.

Unlike mutual information, which lacks a notion of geometry in the feature space and can be difficult to estimate directly, HSIC incorporates geometry via kernel choice and can be directly estimated from mini-batches without restrictive data assumptions \cite{HSIC_SSL}. %
For two random variables $X$ and $Y$, HSIC measures the squared Hilbert-Schmidt norm of the cross-covariance operator between their non-linear feature mappings in reproducing kernel Hilbert spaces (RKHS):
$$
\text{HSIC}(X,Y) = \left\| \mathbb{E}[\phi(X)\otimes\psi(Y)] - \mathbb{E}[\phi(X)] \otimes \mathbb{E}[\psi(Y)] \right\|_{\text{HS}}^2
$$
where $\phi$ and $\psi$ are the feature transformations induced by the chosen kernels for $X$ and $Y$. Crucially, for a wide range of characteristic kernels, $\text{HSIC}(X,Y) = 0$ if and only if $X$ and $Y$ are strictly independent.

Let $Z$ denote the dense patch features outputted by the DINOv3 backbone for the global views.
For each corresponding patch, let $V = (\theta, r_s)$ represent the physical viewing geometry, where $\theta$ is the incidence angle and $r_s$ is the slant range; both are obtained from the recorded acquisition geometry and navigation data.
To incentivize the model to output features that rely solely on the intrinsic reflectivity and structure of the seabed, we want to minimize the statistical dependence between the patch features $Z$ and the viewing geometry $V$.
We can achieve this by introducing an HSIC-based regularization loss during training:
$$
\mathcal{L}_{\text{view-inv}} = \text{HSIC}(Z, V)
$$

Minimizing this term penalizes the network whenever its feature distribution covaries with the incidence angle or the slant range.
The objective acts as an information bottleneck that filters out the geometric and attenuation artifacts specific to a given sonar pass.
It integrates directly into the DINOv3 pretraining objective, yielding the overall loss:
$$
\mathcal{L}_{\text{Total}} = \mathcal{L}_{\text{Pre}} + \gamma \text{HSIC}(Z, V)
$$
where $\mathcal{L}_{\text{Pre}}$ represents the combination of the global discriminative loss ($\mathcal{L}_{\text{DINO}}$), masked latent reconstruction ($\mathcal{L}_{\text{iBOT}}$), and other regularizers, while $\gamma$ is a hyper-parameter controlling the strength of the view-invariance constraint which was set to $0.3$ during training.

However, a naive empirical estimation of HSIC requires computing exact kernel matrices over the mini-batch, resulting in a computational complexity of $\mathcal{O}(B^2)$, where $B$ is the batch size \cite{HSIC_SSL}.
Given that SSS self-distillation often operates under small-batch distributed training constraints, this quadratic complexity can become a bottleneck.
To maintain training efficiency, we utilize Random Fourier Features (RFF) \cite{rahimi2007rff} to approximate a Gaussian Radial Basis Function (RBF) kernel.
By projecting the data into a randomized low-dimensional Fourier space before computing the covariance, the computational complexity of the HSIC regularizer is reduced to $\mathcal{O}(B)$, scaling linearly with the batch size \cite{rahimi2007rff}.
Furthermore, to ensure stable gradients, the bandwidth of the Gaussian RBF kernel is dynamically calibrated during each forward pass using the median distance heuristic of the batch.
This implementation ensures that enforcing strict statistical view-invariance adds minimal computational overhead to the DINOv3 training pipeline.

\section{Experimental Setup}

\subsection{Dataset and Preprocessing}
\label{sec:dataset}

We utilize the BenthiCat dataset, a large-scale opti-acoustic dataset for benthic classification and habitat mapping \cite{rajani2025}, to train and evaluate our framework. 
The dataset contains a subset comprising approximately one million SSS tiles for self-supervised representation learning, derived from data collected along the entire coast of Catalonia, Spain, covering a wide variety of benthic habitats. 
These original SSS data were acquired by Tecnoambiente SLU through a series of geophysical surveys extending up to six nautical miles offshore, spanning more than $1{,}900$~km$^2$ of seabed over $3{,}500$+ hours of acquisition. 
The surveyed area was subdivided into 22 sectors (16 along the northern and 6 along the southern coast), each comprising multiple overlapping transects that together yielded over 50 million swaths.

The surveys were conducted from the 16~m survey vessel \textit{Festina Lente}, which towed a Klein 3000H dual-frequency (500/900~kHz) side-scan sonar. 
The towfish was hauled behind the vessel by a cable controlling its operating depth, flying at an altitude of 5--15~m above the seabed (approximately 10\% of the sensor range). 
The range per channel was manually adjusted for each zone depending on the water depth, varying between 50 and 150~m for the port and starboard channels. 
Only the 500~kHz data were utilized, as they offer the best balance between along-track resolution and area coverage per transect. 
Furthermore, onboard navigation was aided by an inertial navigation system and an ultrashort baseline (USBL) transceiver, providing recursive corrections to navigation errors and online georeferencing of the swaths.

The raw SSS waterfall data then underwent rigorous preprocessing to ensure stable feature learning.
To preserve low-intensity structural details while mitigating extreme outliers, the raw intensities were logarithmically compressed, clipped at the 90th percentile, and normalized to the range $[0,1]$.
Rather than relying on a flat-seafloor assumption for slant-range correction, the acoustic blind zone and nadir noise were explicitly excluded by dropping the central bins.
The remaining port and starboard records were extracted into $768 \times 768$ pixel patches with a 75\% overlap, paired with normalized cross-track distance encodings.
During the self-supervised training pipeline, these patches undergo the previously discussed domain-specific augmentations before being processed by our multi-crop strategy to produce global crops of $224 \times 224$ pixels and local crops of $96 \times 96$ pixels.

\subsection{Training Details}
\label{sec:training}

Direct application of DINO-style self-distillation to SSS imagery is inherently challenging due to acoustic-specific statistics (e.g., speckle noise and gain variations) and the small-batch distributed training constraints typical in this domain. We introduce two adaptations that stabilize training under these conditions. 

First, because the original DINOv3 high-dimensional output ($K=65,536$) produces near-zero gradients at smaller batch sizes, we drastically reduced the projection-head dimensionality to $K=4,096$. Second, the standard Sinkhorn-Knopp algorithm frequently produces zero rows or columns under batch constraints, leading to numerical instability. Therefore, for the global-to-local distillation loss, we reverted to the DINOv1 centering mechanism~\cite{caron2021}: 
\begin{equation}
C_{t} \leftarrow m\, C_{t-1} + (1-m)\, \frac{1}{B} \sum_{i=1}^{B} O_{\text{teacher}}(x_i)
\end{equation}
where the teacher output center $C$ is subtracted prior to the softmax activation. The iBOT masked-image-modeling head retains Sinkhorn-Knopp, as its dense patch tokens provide sufficient statistical density to prevent collapse.

The network was trained across two NVIDIA Quadro RTX 6000 GPUs with an
effective batch size of 8{,}192 ($4{,}096$ per GPU $\times$ 2).
Optimization was performed with AdamW \cite{adamw} using a peak learning rate of
$1.41 \times 10^{-4}$, obtained by scaling a base rate of
$5 \times 10^{-5}$ with the square root of the effective batch size
relative to a reference of $1{,}024$.
The learning rate was linearly increased from $0$ to the peak value over
the first $10$ epochs ($10\%$ of training), then decayed according to a
cosine schedule to a final value of $1 \times 10^{-6}$ over the remaining
$90$ epochs.

\subsection{Unsupervised Feature Space Evaluation}
Given the absence of standardized benchmarks to explicitly measure view-invariance in SSS imagery, we establish a suite of self-supervised proxy metrics. To rigorously verify that the framework successfully isolates intrinsic seabed structures from transient acoustic acquisition geometries\textemdash without relying on downstream task biases\textemdash we comprehensively analyze the geometry, topology, and contrastive dynamics of the fused 768-dimensional feature space ($\mathbb{R}^{768}$).

\subsubsection{Representation Capacity and Uniformity}
A primary failure mode of masked self-supervised learning is dimensional collapse, where the network maps all inputs to a low-dimensional subspace, severely bottlenecking its representational capacity \cite{svd}. To ensure the network fully utilizes the ambient space, we evaluate the feature covariance matrix via Singular Value Decomposition (SVD). 
For a mean-centered feature matrix $\mathbf{X} \in \mathbb{R}^{N \times D}$, we decompose $\mathbf{X} = \mathbf{U} \mathbf{\Sigma} \mathbf{V}^T$ \cite{svd}. The effective rank is computed using the Shannon entropy of the normalized singular values $\sigma_i$, quantifying the uniformity of the variance distribution across latent dimensions. 

To complement the linear SVD analysis, we evaluate the contrastive \textit{Uniformity} score \cite{alignment_uniformity}. This metric measures how evenly the feature vectors are distributed across the unit hypersphere. A high uniformity score confirms that the model avoids crowding representations into a narrow, homogeneous acoustic domain, preserving maximum information entropy to separate distinct benthic textures.

\begin{figure}[!t]
    \centering
    \includegraphics[width=\columnwidth]{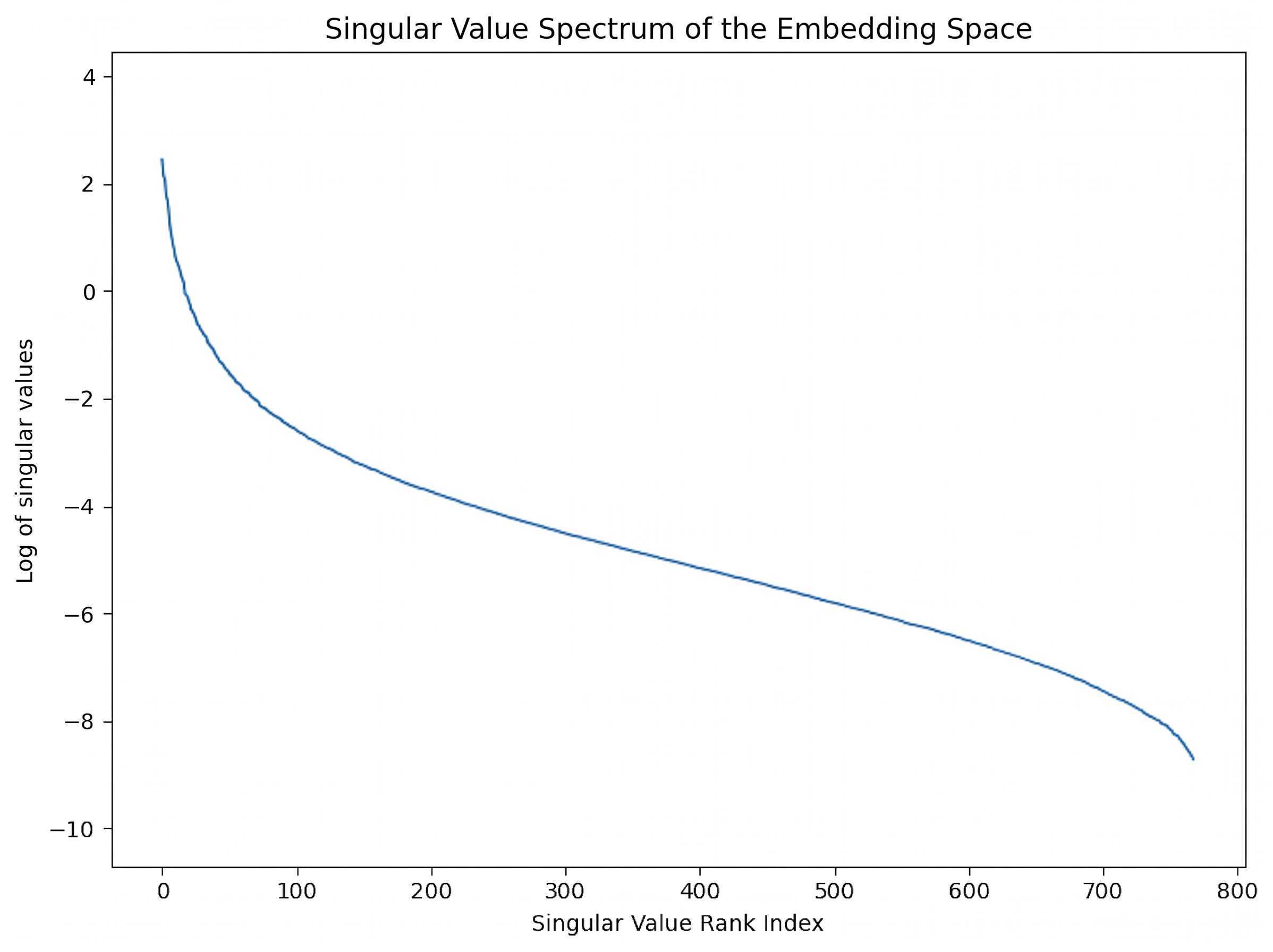}
    \caption{The log of singular values against the singular value rank index of the fused representation.}
    \label{fig:svd}
\end{figure}

\subsubsection{Invariance Probing and Alignment}
To explicitly quantify the model's robustness to transient acoustic dynamics, we measure feature invariance across our physics-guided augmentations. We utilize the contrastive \textit{Alignment} score \cite{alignment_uniformity} alongside direct cosine similarity invariance probing. These metrics calculate the expected distance between heavily perturbed, augmented views of the exact same physical seabed patch. By tracking the mean, minimum, and standard deviation of these similarities, we can directly measure the success of the HSIC penalty in scrubbing spatial attenuation, speckle, and incidence angle artifacts from the latent representations.

\subsubsection{Intrinsic Dimension}
We estimate the intrinsic dimension of the learned manifold using the Two-NN algorithm~\cite{two_nn}. 
This method relies on the ratio of distances to the first and second nearest neighbors to estimate the minimal number of parameters required to describe the local data distribution, free from global Euclidean assumptions. A low intrinsic dimension relative to a high effective rank indicates that while the global space is expressive, the local semantic neighborhoods have successfully compressed and discarded high-variance nuisance parameters (e.g., viewing geometry).

\subsubsection{Manifold Topology and Clustering}
To evaluate the separability and semantic grouping of the unannotated representations, we apply both K-Means \cite{kmeans} and HDBSCAN \cite{hdbscan}. 
Clustering quality is initially quantified using the Silhouette Coefficient~\cite{silhouette}, defined for a single sample as:
\begin{equation}
s = \frac{b - a}{\max(a, b)}
\end{equation}
where $a$ is the mean intra-cluster distance and $b$ is the mean nearest-cluster distance. Values approaching $1$ indicate dense, well-separated spherical clusters in Euclidean space. 

Because continuous acoustic sediment transitions often form complex, non-linear density basins rather than simple isotropic spheres, we apply HDBSCAN post-UMAP projection~\cite{umap}. For these non-spherical density estimations, we strictly track the \textit{Noise Ratio} (the percentage of unclustered, diffuse samples) and the \textit{Mean Confidence} (the probability of strict cluster membership) \cite{hdbscan}. 
Finally, we measure \textit{Trustworthiness} \cite{trustworthiness} to ensure that the local topological neighborhoods of the high-dimensional feature space are faithfully preserved during projection and clustering, heavily penalizing the false preservation of distant points as nearest neighbors.

\section{Results and Discussion}

\subsection{Unsupervised Feature Space Evaluation}
To understand the geometric and semantic properties of the learned representations prior to any downstream supervised fine-tuning, we systematically evaluated the 768-dimensional fused feature space. By probing the latent geometry, we can explicitly quantify how well the framework isolates intrinsic benthic structures from transient acoustic acquisition dynamics.

\subsubsection{Representation Capacity and Uniformity}
A critical failure mode in self-supervised masked modeling is dimensional collapse, where the network projects inputs into a trivial low-dimensional subspace, limiting representation capacity. Our evaluation of the feature covariance matrix yielded an effective rank of $53.8$. While lower than the ambient dimension of $768$, the log-singular value spectrum exhibits a smooth, continuous decay rather than a sharp cutoff (Fig.~\ref{fig:svd}). 

To probe this further, we evaluated the contrastive Uniformity score of the latent space \cite{alignment_uniformity}, which measures how well the feature vectors are distributed across the unit hypersphere. The network achieved a Uniformity score of $-3.13$, confirming that the ConvNeXt-v2 backbone successfully leverages a highly expressive, non-collapsed subspace. Rather than crowding into a narrow acoustic domain, the features maintain maximal information entropy, ensuring that distinct seabed textures remain separable.

Furthermore, the Two-NN algorithm estimated the intrinsic dimension of the data manifold to be $8.80$. This indicates that while the global features span a robust subspace, the local semantic neighborhoods are governed by approximately 9 degrees of freedom. This strong local compression suggests the model successfully filtered out high-variance, viewpoint-dependent acoustic noise (e.g., speckle and transient geometry) to isolate the underlying intrinsic benthic structures.

\begin{figure}[!t]
    \centering
    \includegraphics[width=\columnwidth]{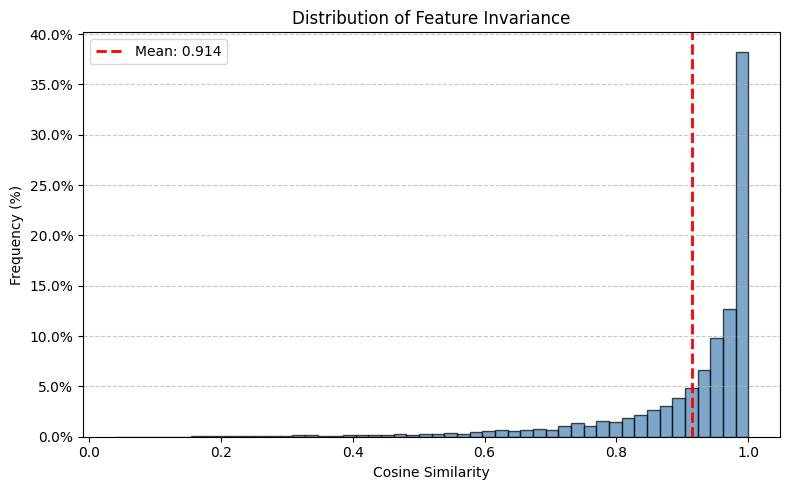}
    \caption{Distribution of feature invariance across augmented pairs. The dashed line indicates a mean cosine similarity of $0.914$, with half of the data points falling below this line and the other half densely concentrated above it.}
    \label{fig:feature_invariance}
\end{figure}

\subsubsection{Quantifying View-Invariance and Alignment}
The core objective of our physics-guided framework is to enforce strict view-invariance. To measure the success of the HSIC penalty and physical augmentations at scrubbing geometric acquisition data, we conducted an invariance ablation study across augmented views. 

The latent space achieved an Alignment score of $0.898$ \cite{alignment_uniformity}, demonstrating that the network consistently maps severely perturbed views of the same physical seabed patch to nearly identical coordinates in the feature space. This is further supported by our direct invariance probing, which yielded a mean cosine similarity of $0.914$ (with a standard deviation of $0.128$) across augmented pairs. As illustrated in Fig.~\ref{fig:feature_invariance}, the scores are heavily skewed toward absolute similarity, with nearly 40\% of augmented pairs achieving a cosine similarity of $1.0$. This directly validates our methodology: by penalizing statistical dependence on incidence angle and slant range, the model natively decouples intrinsic reflectivity from transient viewing conditions.

\begin{figure}[!t]
    \centering
    \includegraphics[width=\columnwidth]{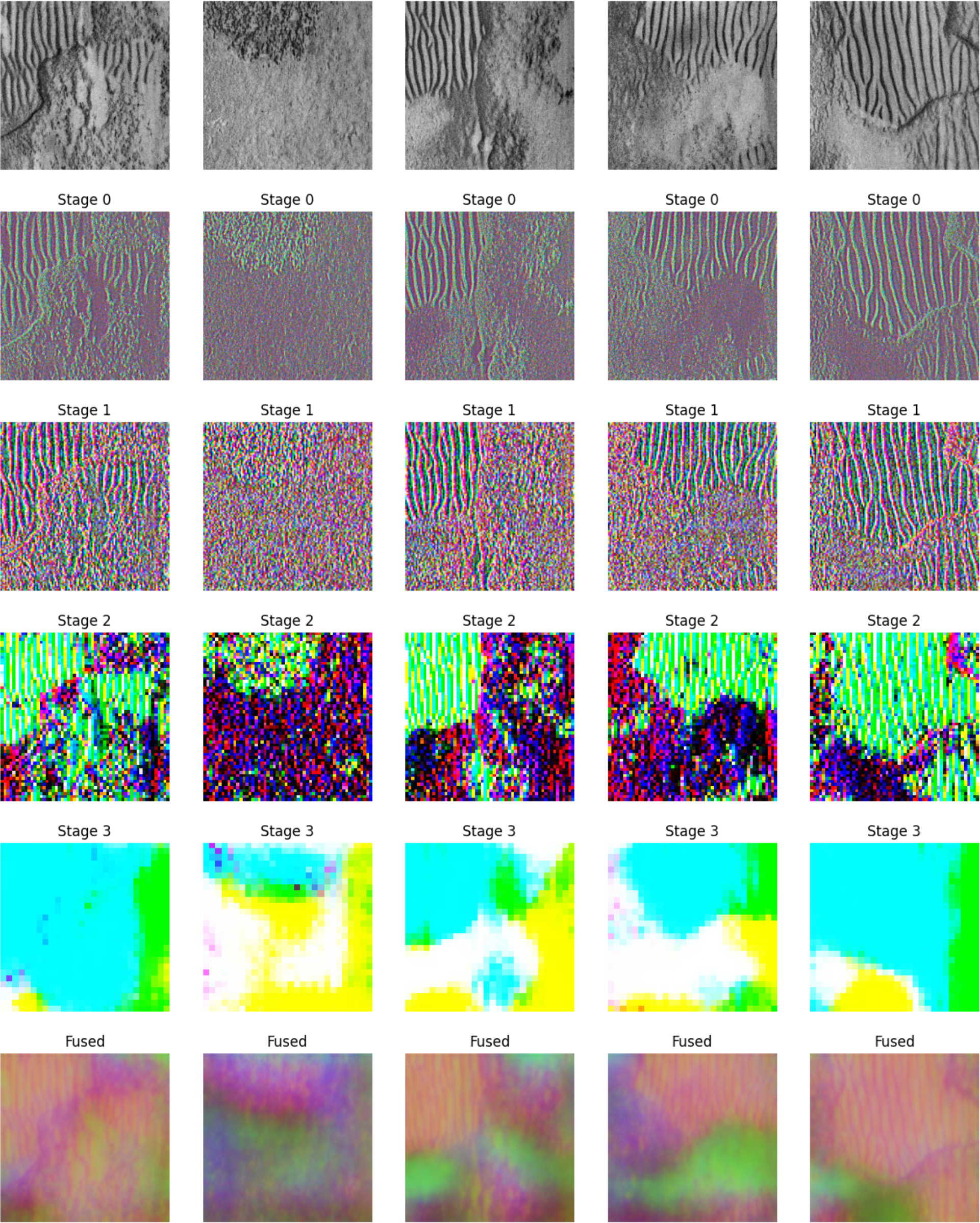}
    \caption{PCA Visualizations of feature maps across network stages extracted from Synthetic Aperature Sonar (SAS) images from the SASSED dataset \cite{cobb2022sassed}.}
    \label{fig:sas}
\end{figure}

\subsubsection{Manifold Geometry and Semantic Clustering}
We assessed the structural grouping of the unannotated representations using both K-Means and HDBSCAN. A K-Means sweep across $k \in [4, 64]$ revealed consistently high Trustworthiness, rapidly scaling to $1.0000$ for $k \geq 32$. This indicates that the local topological neighborhoods are perfectly preserved in the latent space. However, the K-Means Silhouette score remained relatively modest, peaking at $0.134$ for $k=56$. 

This discrepancy implies that the feature space is not partitioned into simple, spherical Euclidean clusters. Because acoustic textures transition continuously based on sediment composition, the underlying manifold is complex. When applying HDBSCAN\textemdash which is capable of discovering non-linear, arbitrarily shaped density basins \cite{hdbscan}\textemdash clustering performance improved dramatically. Upon projecting the space to its intrinsic dimension, HDBSCAN identified 6 distinct, highly dense clusters with an impressive filtered Silhouette score of $0.536$. 

Most notably, the algorithm reported a $0.0\%$ noise ratio alongside a mean cluster membership confidence of $0.995$. There are virtually no diffuse or ambiguous samples bridging the gaps between these semantic basins. This proves that the combination of high feature uniformity, strong positive alignment, and the HSIC objective successfully forces the representations into highly stable, discriminative textures completely unsupervised.

\subsection{Supervised Downstream Evaluation on S3Seg}

\begin{table*}[t]
\centering
\caption{Few-Shot Generalization: Linear Probe vs. K-NN using the Fused Representation}
\label{tab:few_shot}
\begin{tabular}{llccccc}
\toprule
\textbf{Evaluator} & \textbf{Metric} & \textbf{1\% Data} & \textbf{5\% Data} & \textbf{10\% Data} & \textbf{50\% Data} & \textbf{100\% Data} \\
\midrule
\multirow{3}{*}{\textbf{Linear Probe}}
& Accuracy & 0.860 & 0.919 & 0.930 & 0.946 & 0.949 \\
& Macro F1 & 0.825 & 0.897 & 0.914 & 0.932 & 0.936 \\
& mIoU     & 0.706 & 0.817 & 0.844 & 0.876 & 0.882 \\
\midrule
\multirow{3}{*}{\textbf{K-NN}}
& Accuracy & 0.841 & 0.886 & 0.904 & 0.927 & 0.935 \\
& Macro F1 & 0.762 & 0.836 & 0.865 & 0.899 & 0.912 \\
& mIoU     & 0.632 & 0.727 & 0.769 & 0.821 & 0.842 \\
\bottomrule
\end{tabular}
\end{table*}

To evaluate the semantic discriminative power of our self-supervised representations, we conduct extensive supervised ablation studies on the S3Seg dataset \cite{s3seg}, which provides pixel-wise annotations for four distinct benthic classes: Sand Ripples, Fine Sediments, Rocks, and Maerl. We freeze the pretrained ConvNeXt-v2 backbone and evaluate the representations using two standard protocols: Linear Probing (LP) \cite{linear_probe} and K-Nearest Neighbors (K-NN) \cite{knn}.

To rigorously assess the data-efficiency of our model, we simulate a few-shot learning environment by strictly limiting the amount of annotated training data available to the evaluators, sweeping across $1\%$, $5\%$, $10\%$, $50\%$, and $100\%$ subsets of the training split.

\subsubsection{Feature Representation and Stage-by-Stage Progression}
A primary focus of our ablation study is tracking the evolution of the feature representations through the network to justify our multi-scale feature fusion. Table~\ref{tab:stage_progression} presents the Accuracy, Macro F1, and Mean Intersection over Union (mIoU) for features extracted at individual backbone stages compared against our final layer fusion module (the Fused Representation).

\begin{table}[!t]
\centering
\caption{Stage-by-Stage Feature Progression (100\% Data, Linear Probe)}
\label{tab:stage_progression}
\begin{tabular}{lccc}
\toprule
\textbf{Representation} & \textbf{Accuracy} & \textbf{Macro F1} & \textbf{mIoU} \\
\midrule
Stage 1 & 0.686 & 0.578 & 0.434 \\
Stage 2 & 0.815 & 0.757 & 0.627 \\
Stage 3 & 0.938 & 0.920 & 0.855 \\
Stage 4 & 0.932 & 0.917 & 0.849 \\
\midrule
\textbf{Fused Representation} & \textbf{0.949} & \textbf{0.936} & \textbf{0.882} \\
\bottomrule
\end{tabular}
\end{table}

Using the full dataset Linear Probe as a baseline, the quality of the representations improves dramatically through the network. Stage 1 captures basic textures but struggles with semantic boundaries ($0.434$ mIoU). Stage 2 exhibits a strong $+19.3\%$ jump to $0.627$ mIoU as low-level features, such as acoustic shadows and highlights, begin to group. Stage 3 provides another massive $+22.8\%$ improvement ($0.855$ mIoU). However, performance plateaus and slightly dips at Stage 4 ($0.849$ mIoU). This plateau is expected in heavily downsampled convolutional networks, where intermediate layers lose fine-grained spatial details before the fusion layers solidify global concepts while restoring granularity.

Our Fused Representation strictly dominates, peaking at \textbf{0.882} mIoU and \textbf{0.949} accuracy. By preserving the fine-grained high-frequency acoustic details from earlier stages alongside the deep semantic context from later stages, the final fusion layer successfully consolidates the features for optimal class separation.

\subsubsection{Classifier Generalization: Linear Probe vs. K-NN}
Our results, detailed in Table~\ref{tab:few_shot}, compare the generalization capabilities of LP and K-NN across varying data scales. While both evaluators achieve strong peak performance at the 100\% scale ($0.882$ and $0.842$ mIoU, respectively), Linear Probing demonstrates a consistent advantage, particularly in highly constrained data regimes. At the extreme 1\% few-shot scale, LP outperforms K-NN by a notable margin of $0.074$ mIoU.

The performance gap stems from the classifiers' distinct inductive biases in the embedding space. As a non-parametric, local rule, K-NN depends on the density of labeled anchors \cite{knn, silhouette}: with sparse data (1\%--10\%), local distance metrics are brittle and sensitive to noisy acoustic micro-textures, but as data grows to 50\%--100\%, the space becomes dense enough for K-NN to rapidly close the gap. Conversely, the Linear Probe learns a global hyperplane, whose parametric form acts as an inherent regularizer against localized high-frequency noise, favoring linearly separable semantic concepts. This makes LP highly data-efficient: exploiting the strong native separability of the pretrained embeddings, it generalizes well to unseen sonar swaths even from very few annotated samples.

\begin{table}[!t]
\centering
\caption{Per-Class Performance for the Fused Representation (100\% Data, Linear Probe)}
\label{tab:per_class}
\begin{tabular}{lcccc}
\toprule
\textbf{Benthic Class} & \textbf{Precision} & \textbf{Recall} & \textbf{F1-Score} & \textbf{IoU} \\
\midrule
Sand Ripples    & 0.960 & 0.957 & 0.959 & 0.920 \\
Fine Sediments  & 0.984 & 0.974 & 0.979 & 0.959 \\
Rocks           & 0.847 & 0.893 & 0.869 & 0.769 \\
Maerl           & 0.948 & 0.925 & 0.937 & 0.881 \\
\bottomrule
\end{tabular}
\end{table}

\subsubsection{Data Sample Efficiency}
One of the primary promises of the DINO methodology is extreme sample efficiency, which our model delivers. Looking at how the Linear Probe Test mIoU scales for our Fused Representation in Table~\ref{tab:few_shot}, we observe steady, rapid growth from $1\%$ to $10\%$ data scale. 

Remarkably, by utilizing just $10\%$ of the labeled data, the model achieves nearly $96\%$ of its absolute peak performance ($0.844$ vs $0.882$ mIoU). Furthermore, scaling from $50\%$ to $100\%$ yields a marginal $+0.6\%$ gain in mIoU. This confirms the success of our unsupervised pretraining: the learned representations are natively well-separated in the latent space, requiring very few supervised labels to draw accurate classification boundaries. Consequently, the exceptional data-efficiency of this model makes it an excellent candidate for downstream semi-supervised and supervised fine-tuning.

\subsubsection{Class-Level Bottleneck Analysis}
To understand the remaining error modes, we conducted a class-level deep dive on our best configuration (Fused Representation, 100\% Data, Linear Probe). The per-class metrics are reported in Table~\ref{tab:per_class}.

The \textit{Rocks} class serves as the primary bottleneck dragging down the overall dataset mIoU. Analysis of the normalized confusion matrix reveals that true Rocks are misclassified as \textit{Sand Ripples} $8.8\%$ of the time, and as \textit{Maerl} $1.6\%$ of the time. 

We attribute this to two primary factors. First, acoustic similarity: in SSS imagery, both rocks and sand ripples create distinct high-intensity returns followed by hard acoustic shadows. Given the spatial resolution constraints of the extracted patch, the input data sometimes fails to capture the broader, repetitive structural pattern of a ripple field versus the isolated geometric nature of a rock, causing the embedding space to confuse them. Second, class imbalance: the support size for Rocks is significantly smaller than for Sand Ripples or Fine Sediments. The linear probe is therefore slightly biased toward predicting the majority classes to minimize global cross-entropy loss.

\subsection{Cross-Domain Generalization: Shipwreck Segmentation on AI4Shipwrecks}
\label{sec:ai4shipwrecks}

\begin{table}[!t]
\centering
\caption{Site-averaged shipwreck segmentation performance on the AI4Shipwrecks test split.}
\label{tab:ai4_aggregate}
\begin{tabular}{lcccc}
\toprule
\textbf{Method} & \textbf{IoU$_{\text{ship}}$} & \textbf{F1} & \textbf{TPR} & \textbf{TNR} \\
\midrule
Yang \textit{et al.} \cite{yang2022sss}   & 0.212 & 0.310 & 0.296 & 0.995 \\
ViT-Adapter \cite{chen2023vitadapter}     & 0.283 & 0.395 & 0.488 & 0.995 \\
DeepLabv3 \cite{chen2017deeplab}          & 0.363 & 0.473 & 0.485 & 0.996 \\
HRNet \cite{wang2020hrnet}                & 0.372 & 0.490 & 0.516 & 0.998 \\
U-Net \cite{ronneberger2015unet}          & 0.411 & 0.526 & 0.592 & 0.997 \\
SOD (InSPyReNet) \cite{kim2022inspyre}    & \textbf{0.445} & \textbf{0.594} & \textbf{0.652} & \textbf{0.997} \\
\midrule
\textbf{Ours (frozen encoder + U-Net)}    & 0.423 & 0.553 & 0.535 & 0.988 \\
\bottomrule
\end{tabular}
\end{table}

To assess whether the learned representations transfer beyond benthic habitat mapping, we evaluate them on the AI4Shipwrecks benchmark \cite{ai4shipwrecks}, a pixel-wise annotated side-scan sonar dataset of 28 shipwreck sites surveyed by an Iver3 autonomous underwater vehicle equipped with an EdgeTech 2205 dual-frequency (520\,kHz CHIRP) SSS in Thunder Bay National Marine Sanctuary (Lake Huron, MI, USA). This benchmark is a deliberately adversarial transfer setting for our framework, imposing a triple domain gap along the following axes:

\begin{enumerate}
    \item \textbf{Sensor and environment mismatch.} The pretraining corpus (BenthiCat) was acquired along the Catalan coast of the Mediterranean Sea using a surface-towed Klein 3000H (500\,kHz). AI4Shipwrecks originates from a cold, fresh-water inland lake, was recorded using an AUV platform (introducing different hydrodynamic stability and navigation profiles) equipped with an EdgeTech 2205 CHIRP system, and spans survey geometries (1--45\,m altitude, 30--150\,m ranges) and resolutions (13.6--72.6\,cm along-track) distinct from the pretraining data.
    \item \textbf{Radiometric and geometric distribution mismatch.} During pretraining, BenthiCat tiles were logarithmically compressed, clipped, and had the acoustic blind zone (nadir) completely removed (Sec.~\ref{sec:dataset}). Conversely, AI4Shipwrecks imagery is linearly normalized, TVG-compensated by the acquisition software, and retains the full uncorrected water column. The encoder is therefore forced to process intensity statistics and nadir geometries that it never observed during self-supervised training.
    \item \textbf{Out-of-class targets.} The pretraining corpus consists exclusively of natural seabed; it contains little to no man-made structures of any kind, and shipwrecks in particular are entirely absent. All object-level discrimination must therefore be learned by the downstream decoder from the benchmark's annotations alone.
\end{enumerate}

\subsubsection{Experimental Protocol}
\label{sec:ai4_protocol}
We follow the benchmark's official 50/50, site-disjoint train--test split, which guarantees that no survey of a given wreck appears in both sets. The pretrained encoder is kept entirely frozen, and the fused 768-dimensional dense features (Sec.~\ref{sec:encoder}) are fed to a standard U-Net decoder \cite{ronneberger2015unet} trained from scratch on the labeled training split; shipwreck pixels constitute only $2.9\%$ of the test set. In line with the benchmark's evaluation protocol, we report the site-averaged Intersection over Union of the shipwreck class (IoU$_{\text{ship}}$), F1-score, True Positive Rate (TPR), and True Negative Rate (TNR), computed over the 13 wreck-bearing test sites. \cite{ai4shipwrecks}.

\begin{table}[!t]
\centering
\caption{Per-site IoU$_{\text{ship}}$ on the 13 wreck-bearing test sites, sorted by depth in descending order.}
\label{tab:ai4_persite}
\small
\begin{tabular}{lccccc}
\toprule
\textbf{Site} & \textbf{Depth} & \textbf{Ours} & \textbf{Best baseline} & \textbf{Best} \\
 & (m) & & (model) & \\
\midrule
W.H. Gilbert        & 77.7 & 0.655 & SOD        & 0.749 \\
Viator              & 57.3 & 0.729 & SOD        & 0.776 \\
Corsair             & 55.5 & 0.694 & SOD        & 0.583 \\
Corsican            & 48.8 & 0.566 & HRNet      & 0.174 \\
Monrovia            & 42.7 & 0.612 & U-Net      & 0.572 \\
W.P. Thew           & 25.6 & 0.405 & SOD        & 0.580 \\
L. van Valkenburg   & 18.3 & 0.483 & SOD        & 0.713 \\
Barge No.~1         & 12.8 & 0.426 & SOD        & 0.775 \\
J. Davidson         & 10.7 & 0.002 & U-Net      & 0.074 \\
Monohansett         & 5.5  & 0.075 & ViT-Adapter& 0.496 \\
Haltiner Barge      & 5.2  & 0.435 & DeepLabv3  & 0.459 \\
Shamrock            & 3.4  & 0.210 & U-Net      & 0.428 \\
Artificial Reef     & 3.0  & 0.201 & SOD        & 0.017 \\
\bottomrule
\end{tabular}
\end{table}

\subsubsection{Aggregate Performance}
\label{sec:ai4_aggregate}

Table~\ref{tab:ai4_aggregate} reports the site-averaged comparison against the benchmark's six published baselines. Despite the triple domain gap enumerated above, our frozen representations combined with a lightweight decoder achieve an IoU$_{\text{ship}}$ of $0.423$ and an F1-score of $0.553$, ranking second among the seven models and outperforming \textit{every} segmentation network trained end-to-end from scratch on this dataset, including the benchmark's own U-Net ($0.411$ IoU$_{\text{ship}}$, $0.526$ F1), DeepLabv3 ($0.363$), HRNet ($0.372$), ViT-Adapter ($0.283$), and Yang \textit{et al.} ($0.212$). Only the salient-object-detection model InSPyReNet, which is itself initialized from large-scale \textit{optical} pretraining, remains ahead ($0.445$). We consider this a strong result for the generality of the learned acoustic representations: a model that never observed a single man-made structure, a different sonar payload, and a radiometrically mismatched distribution still outperforms fully supervised models that had full access to the in-domain training imagery and its labels. Notably, the encoder contributes this transfer without a single gradient step on AI4Shipwrecks data; every wreck-specific decision boundary is learned by the decoder alone.

\subsubsection{Site-Level Analysis: Complementary Expertise}
\label{sec:ai4_persite}

The site-level breakdown in Table~\ref{tab:ai4_persite} reveals a structure that the aggregate metrics conceal. Our model surpasses the best-performing baseline on 4 of the 13 sites, but these wins are concentrated precisely where \textit{all} supervised baselines collapse. On the Corsican site, the strongest baseline reaches only $0.174$ IoU$_{\text{ship}}$ while our features support $0.566$\textemdash a $3.3\times$ improvement\textemdash and on Artificial Reef the benchmark's best model attains a near-total failure ($0.017$) where ours reaches $0.201$. Conversely, the largest deficits (Barge No.~1, Monohansett) occur on sites where at least one supervised baseline performs well. The failure modes are therefore largely \textit{disjoint} from those of the benchmark's models: per-site oracle selection between our model and the strongest baseline would yield $0.548$ IoU$_{\text{ship}}$, a gain of more than 10 points over the best single model ($0.445$), and even a two-model per-site maximum with InSPyReNet alone yields $0.525$. This complementarity indicates that the self-supervised features encode an information source orthogonal to what is learnable from this dataset's supervision signal.

\subsubsection{Depth-Stratified Behavior and Pretraining Coverage}
\label{sec:ai4_depth}

\begin{figure}[!t]
    \centering
    \includegraphics[width=\columnwidth]{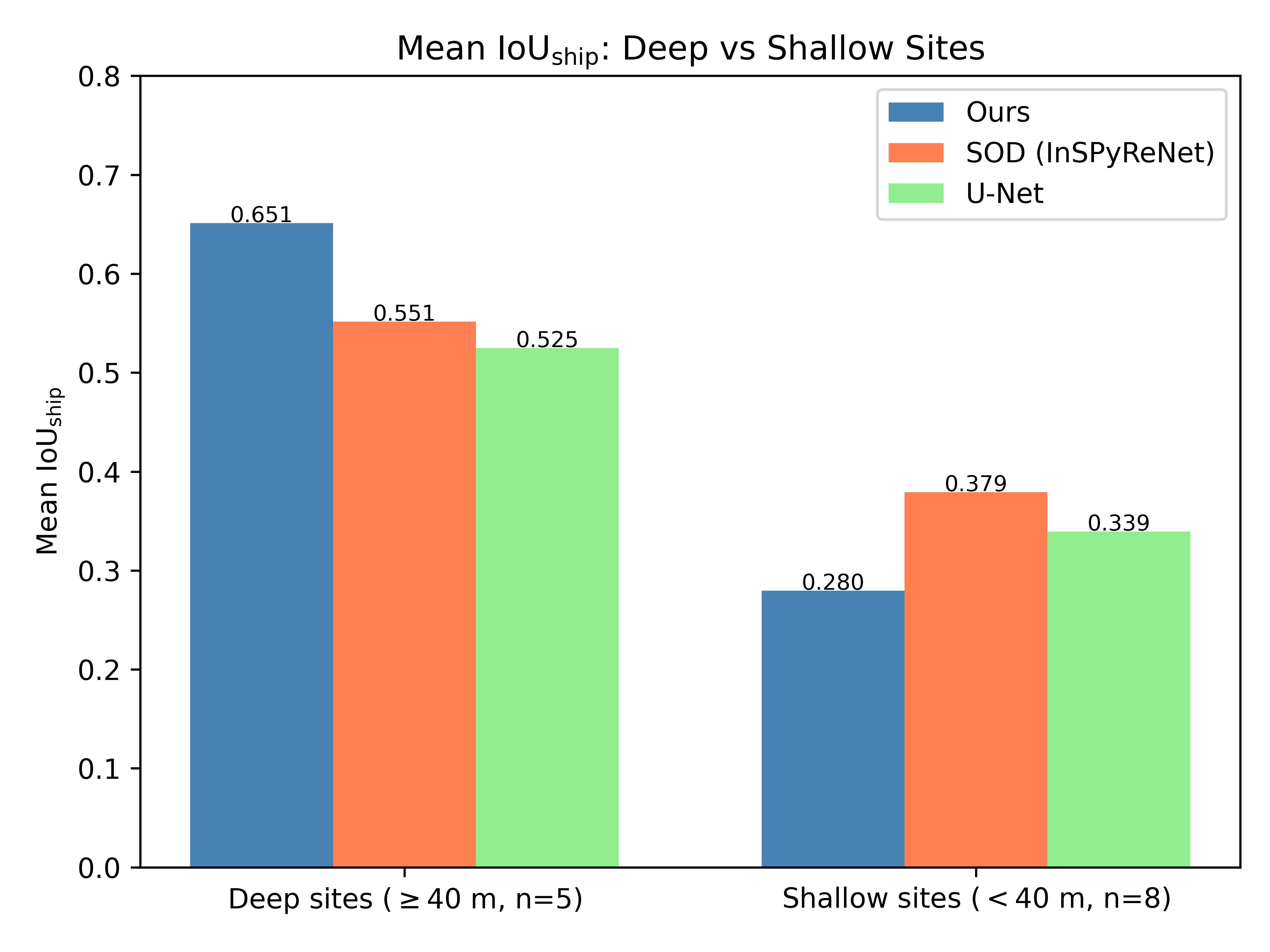}
    \caption{Mean IoU$_{\text{ship}}$ on the AI4Shipwrecks test split, stratified by site depth.}
    \label{fig:depth_bargraph}
\end{figure}

Stratifying the sites by water depth exposes a systematic pattern: the Spearman rank correlation between per-site IoU$_{\text{ship}}$ and site depth is $\rho = 0.83$. On the five sites deeper than 40\,m (Viator, Corsair, Corsican, Monrovia, W.H. Gilbert), our method attains a mean IoU$_{\text{ship}}$ of $0.651$ which is the highest deep-water score of any model in the benchmark, surpassing InSPyReNet ($0.551$) and the benchmark U-Net ($0.525$). On the eight shallow sites ($<40$\,m), performance drops to $0.280$, below the supervised baselines (InSPyReNet: $0.380$, U-Net: $0.340$), as shown in Fig.~\ref{fig:depth_bargraph}.

\begin{figure}[!t]
    \centering
    \includegraphics[width=\columnwidth]{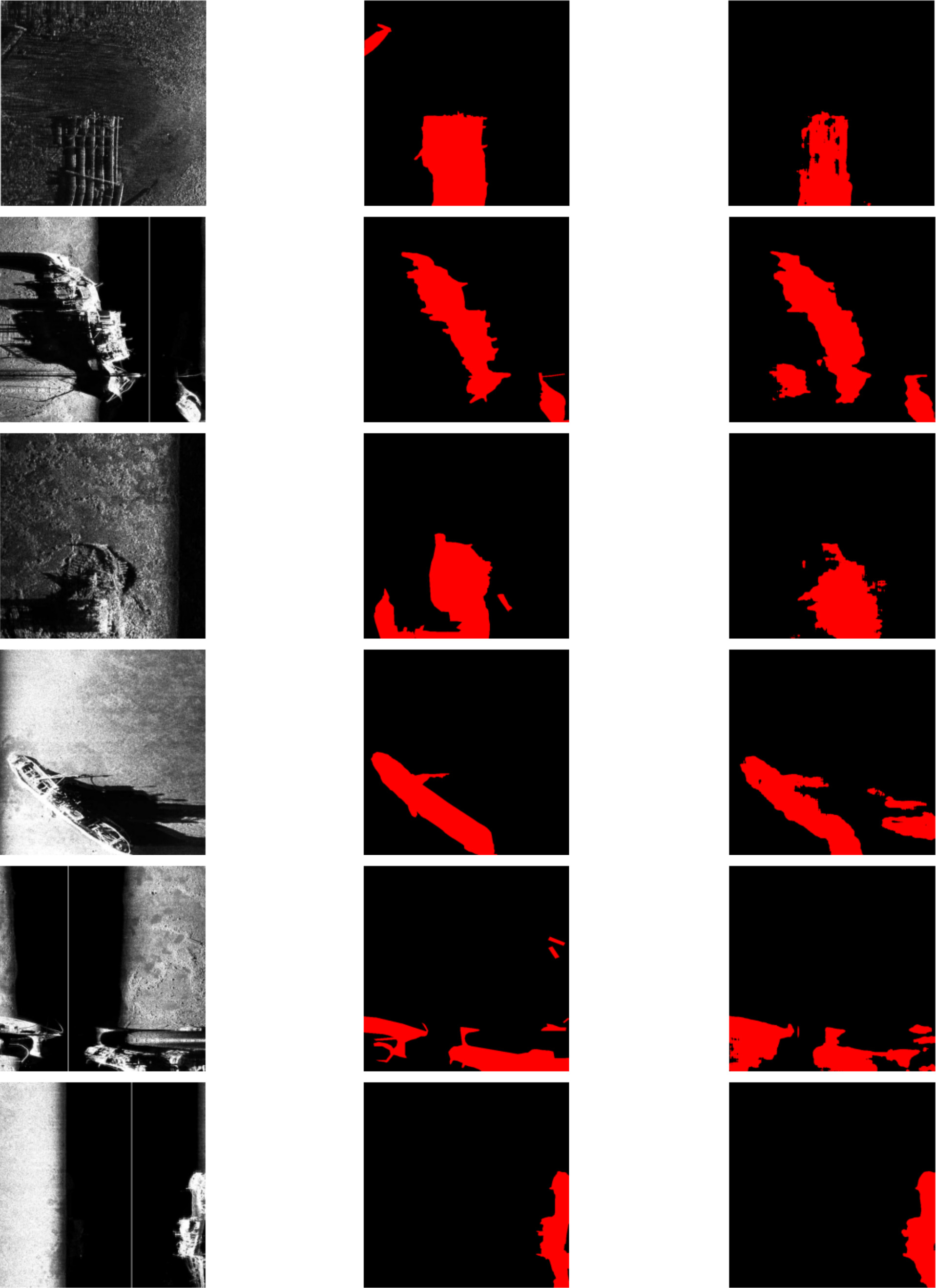}
    \caption{Segmentation results from the BenthicDINO encoder + U-Net head applied to the AI4Shipwrecks dataset \cite{ai4shipwrecks}. Left to right, original SSS image, ground truth annotation, BenthicDINO result.}
    \label{fig:ai4shipwrecks}
\end{figure}

We attribute the observed depth-dependent performance (Fig.~\ref{fig:depth_bargraph}) to the scope of the pretraining corpus rather than any flaw in the representation objective.
The model succeeds when target features align with the pretraining distribution and struggles when they do not:

\begin{itemize}
    \item \textbf{Deep Sites (Strong Transfer):} Imaged at longer ranges and coarser resolutions ($55$--$73$\,cm along-track), deep wrecks appear as compact, high-contrast reflectivity anomalies. This closely matches the low-relief seabed in the BenthiCat pretraining corpus. Consequently, the frozen features generalize better here than any amount of in-domain supervised training.
    \item \textbf{Shallow Sites (Weak Transfer):} Surveyed at low altitudes and short ranges, the wrecks in the shallow AI4Shipwrecks sites project extended acoustic shadows. Because BenthiCat consists entirely of natural seabed lacking prominent vertical structures, these shadow geometries fall completely outside the pretraining distribution, leaving the decoder with little signal to exploit.
\end{itemize}

This indicates that the shallow-site deficit is a data-coverage problem, not a design trade-off. Broadening the pretraining distribution to include shallow, low-grazing-angle surveys with prominent shadow geometry should close this gap without requiring changes to the learning objective itself.

\section{Conclusion and Future Work}

In this work, we introduced BenthicDINO, a physics-informed self-supervised learning framework that overcomes the view-dependence bottleneck in SSS imagery. By integrating a ConvNeXt-v2 backbone, physics-guided acoustic augmentations, and a HSIC penalty, the framework explicitly decouples intrinsic seabed structures from transient acquisition geometries like slant range and incidence angle. 

Our evaluations demonstrate that this approach yields highly robust, data-efficient representations without relying on dense annotations. BenthicDINO achieves near-peak supervised classification performance on benthic habitat mapping using only 10\% of available labels. Furthermore, the frozen representations exhibit exceptional cross-domain generalization, outperforming fully supervised baselines on the AI4Shipwrecks benchmark despite significant sensor, environment, and target-class domain gaps.

Future work will focus on expanding the pretraining corpus to include shallow-water and low-altitude surveys. Exposing the model to more extreme grazing angles and prominent acoustic shadows will address current limitations in shallow-water structural transfer, paving the way for universally robust acoustic perception models across all survey depths and geometries.

\section*{Acknowledgments}

This work was supported by the Spanish Government through the project "Automated Seabed Analysis through Self-Supervised Deep Learning Sonar Technology (ASSiST)" under grant PID2023-149413OB-I00. 

\bibliographystyle{IEEEtran}
\bibliography{bibliography}

@misc{can2025geo_match,
    title={A Geometrically Consistent Matching Framework for Side-Scan Sonar Mapping},
    author={Can Lei and Hayat Rajani and Nuno Gracias and Rafael Garcia and Huigang Wang},
    year={2025},
    eprint={2509.11255},
    archivePrefix={arXiv},
    primaryClass={physics.ins-det},
    url={https://arxiv.org/abs/2509.11255},
}

@article{katrusha2025,
    title={Change detection in side-scan sonar imagery based on deep learning feature matching methods},
    volume={6},
    url={https://journals.uran.ua/eejet/article/view/346940},
    DOI={10.15587/1729-4061.2025.346940},
    number={2 (138)},
    journal={Eastern-European Journal of Enterprise Technologies},
    author={Katrusha, Oleksandr and Prylipko, Dmytro and Yefremov, Kostiantyn},
    year={2025},
    month={Dec.},
    pages={52--62}
}

@article{burguera2016,
    doi={10.1371/journal.pone.0146396},
    author={Burguera, Antoni AND Oliver, Gabriel},
    journal={PLOS ONE},
    publisher={Public Library of Science},
    title={High-Resolution Underwater Mapping Using Side-Scan Sonar},
    year={2016},
    month={01},
    volume={11},
    url={https://doi.org/10.1371/journal.pone.0146396},
    pages={1--41},
    number={1},
}

@misc{rajani2025,
    title={BenthiCat: An opti-acoustic dataset for advancing benthic classification and habitat mapping},
    author={Hayat Rajani and Valerio Franchi and Borja Martinez-Clavel Valles and Raimon Ramos and Rafael Garcia and Nuno Gracias},
    year={2025},
    eprint={2510.04876},
    archivePrefix={arXiv},
    primaryClass={cs.CV},
    url={https://arxiv.org/abs/2510.04876},
}

@article{can2025_physdnet,
  title={PhysDNet: Physics-Guided Decomposition Network of Side-Scan Sonar Imagery},
  author={Lei, Can and Rajani, Hayat and Gracias, Nuno and Garcia, Rafael and Wang, Huigang},
  journal={IEEE Geoscience and Remote Sensing Letters},
  year={2026},
  publisher={IEEE}
}

@article{s3seg,
    title = {A convolutional vision transformer for semantic segmentation of side-scan sonar data},
    journal = {Ocean Engineering},
    volume = {286},
    pages = {115647},
    year = {2023},
    issn = {0029-8018},
    doi = {https://doi.org/10.1016/j.oceaneng.2023.115647},
    url = {https://www.sciencedirect.com/science/article/pii/S0029801823020310},
    author = {Hayat Rajani and Nuno Gracias and Rafael Garcia}
}

@article{coiras2007,
  author={Coiras, Enrique and Petillot, Yvan and Lane, David M.},
  journal={IEEE Transactions on Image Processing}, 
  title={Multiresolution 3-D Reconstruction From Side-Scan Sonar Images}, 
  year={2007},
  volume={16},
  number={2},
  pages={382-390},
  url={https://ieeexplore.ieee.org/document/4060928},
  doi={10.1109/TIP.2006.888337}
}

@inproceedings{alan2022,
  title={Self-supervised learning for sonar image classification},
  author={Preciado-Grijalva, Alan and Wehbe, Bilal and Firvida, Miguel Bande and Valdenegro-Toro, Matias},
  booktitle={2022 IEEE/CVF Conference on Computer Vision and Pattern Recognition Workshops (CVPRW)},
  pages={1498--1507},
  year={2022},
  organization={IEEE}
}

@article{fu2025,
    author={Fu, Yu and Luo, Xiaowen and Qin, Xiaoming and Wan, Hongyang and Cui, Jiaxin and Huang, Zepeng},
    title={Deep Learning-Based Feature Matching Algorithm for Multi-Beam and Side-Scan Images},
    journal={Remote Sensing},
    volume={17},
    year={2025},
    number={4},
    url={https://www.mdpi.com/2072-4292/17/4/675},
    ISSN={2072-4292},
    DOI={10.3390/rs17040675}
}

@misc{oriane2025,
    title={DINOv3},
    author={Oriane Siméoni and Huy V. Vo and Maximilian Seitzer and Federico Baldassarre and Maxime Oquab and Cijo Jose and Vasil Khalidov and Marc Szafraniec and Seungeun Yi and Michaël Ramamonjisoa and Francisco Massa and Daniel Haziza and Luca Wehrstedt and Jianyuan Wang and Timothée Darcet and Théo Moutakanni and Leonel Sentana and Claire Roberts and Andrea Vedaldi and Jamie Tolan and John Brandt and Camille Couprie and Julien Mairal and Hervé Jégou and Patrick Labatut and Piotr Bojanowski},
    year={2025},
    eprint={2508.10104},
    archivePrefix={arXiv},
    primaryClass={cs.CV},
    url={https://arxiv.org/abs/2508.10104},
}

@misc{oquab2024,
    title={DINOv2: Learning Robust Visual Features without Supervision}, 
    author={Maxime Oquab and Timothée Darcet and Théo Moutakanni and Huy Vo and Marc Szafraniec and Vasil Khalidov and Pierre Fernandez and Daniel Haziza and Francisco Massa and Alaaeldin El-Nouby and Mahmoud Assran and Nicolas Ballas and Wojciech Galuba and Russell Howes and Po-Yao Huang and Shang-Wen Li and Ishan Misra and Michael Rabbat and Vasu Sharma and Gabriel Synnaeve and Hu Xu and Hervé Jegou and Julien Mairal and Patrick Labatut and Armand Joulin and Piotr Bojanowski},
    year={2024},
    eprint={2304.07193},
    archivePrefix={arXiv},
    primaryClass={cs.CV},
    url={https://arxiv.org/abs/2304.07193}, 
}

@misc{caron2021,
    title={Emerging Properties in Self-Supervised Vision Transformers}, 
    author={Mathilde Caron and Hugo Touvron and Ishan Misra and Hervé Jégou and Julien Mairal and Piotr Bojanowski and Armand Joulin},
    year={2021},
    eprint={2104.14294},
    archivePrefix={arXiv},
    primaryClass={cs.CV},
    url={https://arxiv.org/abs/2104.14294}, 
}

@inproceedings{woo2023,
  title={Convnext v2: Co-designing and scaling convnets with masked autoencoders},
  author={Woo, Sanghyun and Debnath, Shoubhik and Hu, Ronghang and Chen, Xinlei and Liu, Zhuang and Kweon, In So and Xie, Saining},
  booktitle={2023 IEEE/CVF Conference on Computer Vision and Pattern Recognition (CVPR)},
  pages={16133--16142},
  year={2023},
  organization={Ieee}
}

@inproceedings{liu2022,
  title={A convnet for the 2020s},
  author={Liu, Zhuang and Mao, Hanzi and Wu, Chao-Yuan and Feichtenhofer, Christoph and Darrell, Trevor and Xie, Saining},
  booktitle={2022 IEEE/CVF conference on computer vision and pattern recognition (CVPR)},
  pages={11966--11976},
  year={2022},
  organization={IEEE}
}

@inproceedings{HSIC_SSL,
 author = {Li, Yazhe and Pogodin, Roman and Sutherland, Danica J. and Gretton, Arthur},
 booktitle = {Advances in Neural Information Processing Systems},
 editor = {M. Ranzato and A. Beygelzimer and Y. Dauphin and P.S. Liang and J. Wortman Vaughan},
 pages = {15543--15556},
 publisher = {Curran Associates, Inc.},
 title = {Self-Supervised Learning with Kernel Dependence Maximization},
 url = {https://proceedings.neurips.cc/paper_files/paper/2021/file/83004190b1793d7aa15f8d0d49a13eba-Paper.pdf},
 volume = {34},
 year = {2021}
}

@inproceedings{hsic,
author = {Gretton, Arthur and Bousquet, Olivier and Smola, Alex and Sch"{o}lkopf, Bernhard},
title = {Measuring Statistical Dependence with {Hilbert-Schmidt} Norms},
booktitle = {International Conference on Algorithmic Learning Theory (ALT)},
series = {LNAI},
volume = {3734},
pages = {63--77},
year = {2005},
publisher = {Springer}
}

@inproceedings{rahimi2007rff,
author = {Rahimi, Ali and Recht, Benjamin},
title = {Random Features for Large-Scale Kernel Machines},
booktitle = {Advances in Neural Information Processing Systems (NIPS)},
volume = {20},
pages = {1177--1184},
year = {2007}
}

@article{sift,
author = {Lowe, David G.},
title = {Distinctive Image Features from Scale-Invariant Keypoints},
journal = {International Journal of Computer Vision},
volume = {60},
number = {2},
pages = {91--110},
year = {2004}
}

@article{surf,
author = {Bay, Herbert and Ess, Andreas and Tuytelaars, Tinne and Van Gool, Luc},
title = {Speeded-Up Robust Features ({SURF})},
journal = {Computer Vision and Image Understanding},
volume = {110},
number = {3},
pages = {346--359},
year = {2008}
}

@inproceedings{valdenegro2021pre,
  title={Pre-trained models for sonar images},
  author={Valdenegro-Toro, Matias and Preciado-Grijalva, Alan and Wehbe, Bilal},
  booktitle={OCEANS 2021: San Diego--Porto},
  pages={1--8},
  year={2021},
  organization={IEEE}
}

@inproceedings{sheffield2023advances,
  title={Advances in Self-Supervised Learning for Synthetic Aperture Sonar Data Processing, Classification, and Pattern Recognition},
  author={Sheffield, Brandon W and Bobe, Frank E and Marchand, Bradley and Emigh, Matthew S},
  booktitle={OCEANS 2023-MTS/IEEE US Gulf Coast},
  pages={1--5},
  year={2023},
  organization={IEEE}
}

@article{sheffield2024vit,
  title={On vision transformers for classification tasks in side-scan sonar imagery},
  author={Sheffield, BW and Ellen, Jeffrey and Whitmore, Ben},
  journal={arXiv preprint arXiv:2409.12026},
  year={2024}
}

@article{minejepa2026,
  title={Mine-JEPA: In-Domain Self-Supervised Learning for Mine-Like Object Classification in Side-Scan Sonar},
  author={Kwon, Taeyoun and Choi, Youngwon and Kim, Hyeonyu and Cho, Myeongkyun and Choi, Junhyeok and Kim, Moon Hwan},
  journal={arXiv preprint arXiv:2604.00383},
  year={2026}
}

@article{zhang2025fls,
  title={Self-supervised enhancement of forward-looking sonar images: Bridging cross-modal degradation gaps through feature space transformation and multi-frame fusion},
  author={Zhang, Zhisheng and Zhang, Peng and Wang, Fengxiang and Ma, Liangli and Sun, Fuchun},
  journal={arXiv preprint arXiv:2504.10974},
  year={2025}
}

@inproceedings{kamal2024s3sim,
  title={S3simulator: A benchmarking side scan sonar simulator dataset for underwater image analysis},
  author={Kamal Basha, S and Nambiar, Athira},
  booktitle={International Conference on Pattern Recognition},
  pages={219--235},
  year={2024},
  organization={Springer}
}

@article{acousim2026,
  title={Physics-informed simulation framework for realistic sonar image generation and statistical validation},
  author={Nambiar, Athira and others},
  journal={arXiv preprint arXiv:2605.19712},
  year={2026}
}

@article{li2024side,
  title={Side-scan sonar image generation under zero and few samples for underwater target detection},
  author={Li, Liang and Li, Yiping and Wang, Hailin and Yue, Chenghai and Gao, Peiyan and Wang, Yuliang and Feng, Xisheng},
  journal={Remote Sensing},
  volume={16},
  number={22},
  pages={4134},
  year={2024},
  publisher={MDPI}
}

@article{peng2025multi,
  title={Multi-view sonar image generation via GAN trained with limited data for underwater object classification and detection},
  author={Peng, Ye and Li, Houpu and Zhang, Wenwen and Zhu, Junhui and Liu, Lei and Zhai, Guojun},
  journal={Expert Systems with Applications},
  pages={129452},
  year={2025},
  publisher={Elsevier}
}

@inproceedings{koo2024cyclegan,
  title={Cycle-gan-based synthetic sonar image generation for improved underwater classification},
  author={Koo, Sunmo and Youm, Sangpil and Shin, Jane},
  booktitle={Ocean Sensing and Monitoring XVI},
  volume={13061},
  pages={69--83},
  year={2024},
  organization={SPIE}
}

@article{ma2025diffusion,
  title={Enhancing sonar image segmentation with random fusion in a diffusion model framework: Z. Ma et al.},
  author={Ma, Zhihao and Meng, Weiliang and Zhao, Xixi and Jiang, Longyu},
  journal={The Visual Computer},
  volume={41},
  number={11},
  pages={8369--8383},
  year={2025},
  publisher={Springer}
}

@article{liu2023gray,
  title={A gray scale correction method for side-scan sonar images considering rugged seafloor},
  author={Liu, Yifei and Ye, Xiufen},
  journal={IEEE Transactions on Geoscience and Remote Sensing},
  volume={61},
  pages={1--10},
  year={2023},
  publisher={IEEE}
}

@article{stewart2023height,
  title={Image-to-height domain translation for synthetic aperture sonar},
  author={Stewart, Dylan and Kreulach, Austin and Johnson, Shawn F and Zare, Alina},
  journal={IEEE Transactions on Geoscience and Remote Sensing},
  volume={61},
  pages={1--13},
  year={2023},
  publisher={IEEE}
}

@misc{asifullah2025,
    title={A Survey of the Self Supervised Learning Mechanisms for Vision Transformers},
    author={Asifullah Khan and Anabia Sohail and Mustansar Fiaz and Mehdi Hassan and Tariq Habib Afridi and Sibghat Ullah Marwat and Farzeen Munir and Safdar Ali and Hannan Naseem and Muhammad Zaigham Zaheer and Kamran Ali and Tangina Sultana and Ziaurrehman Tanoli and Naeem Akhter},
    year={2025},
    eprint={2408.17059},
    archivePrefix={arXiv},
    primaryClass={cs.CV},
    url={https://arxiv.org/abs/2408.17059},
}

@misc{adamw,
      title={Decoupled Weight Decay Regularization}, 
      author={Ilya Loshchilov and Frank Hutter},
      year={2019},
      eprint={1711.05101},
      archivePrefix={arXiv},
      primaryClass={cs.LG},
      url={https://arxiv.org/abs/1711.05101}, 
}

@misc{svd,
    title={Understanding Dimensional Collapse in Contrastive Self-supervised Learning}, 
    author={Li Jing and Pascal Vincent and Yann LeCun and Yuandong Tian},
    year={2022},
    eprint={2110.09348},
    archivePrefix={arXiv},
    primaryClass={cs.CV},
    url={https://arxiv.org/abs/2110.09348}, 
}

@article{two_nn,
   title={Estimating the intrinsic dimension of datasets by a minimal neighborhood information},
   volume={7},
   ISSN={2045-2322},
   url={http://dx.doi.org/10.1038/s41598-017-11873-y},
   DOI={10.1038/s41598-017-11873-y},
   number={1},
   journal={Scientific Reports},
   publisher={Springer Science and Business Media LLC},
   author={Facco, Elena and d’Errico, Maria and Rodriguez, Alex and Laio, Alessandro},
   year={2017},
}

@misc{umap,
    title={UMAP: Uniform Manifold Approximation and Projection for Dimension Reduction}, 
    author={Leland McInnes and John Healy and James Melville},
    year={2020},
    eprint={1802.03426},
    archivePrefix={arXiv},
    primaryClass={stat.ML},
    url={https://arxiv.org/abs/1802.03426}, 
}

@InProceedings{hdbscan,
    author="Campello, Ricardo J. G. B.
    and Moulavi, Davoud
    and Sander, Joerg",
    editor="Pei, Jian
    and Tseng, Vincent S.
    and Cao, Longbing
    and Motoda, Hiroshi
    and Xu, Guandong",
    title="Density-Based Clustering Based on Hierarchical Density Estimates",
    booktitle="Advances in Knowledge Discovery and Data Mining",
    year="2013",
    publisher="Springer Berlin Heidelberg",
    address="Berlin, Heidelberg",
    pages="160--172",
}

@article{silhouette,
    title = {Silhouettes: A graphical aid to the interpretation and validation of cluster analysis},
    journal = {Journal of Computational and Applied Mathematics},
    volume = {20},
    pages = {53-65},
    year = {1987},
    issn = {0377-0427},
    doi = {https://doi.org/10.1016/0377-0427(87)90125-7},
    url = {https://www.sciencedirect.com/science/article/pii/0377042787901257},
    author = {Peter J. Rousseeuw}
}

@InProceedings{trustworthiness,
    title = {Learning a Parametric Embedding by Preserving Local Structure},
    author = {van der Maaten, Laurens},
    booktitle = {Proceedings of the Twelfth International Conference on Artificial Intelligence and Statistics},
    pages = {384--391},
    year = {2009},
    editor = {van Dyk, David and Welling, Max},
    volume = {5},
    series = {Proceedings of Machine Learning Research},
    address = {Hilton Clearwater Beach Resort, Clearwater Beach, Florida USA},
    month = {16--18 Apr},
    publisher = {PMLR},
    url = {https://proceedings.mlr.press/v5/maaten09a.html},
}

@misc{alignment_uniformity,
      title={Understanding Contrastive Representation Learning through Alignment and Uniformity on the Hypersphere}, 
      author={Tongzhou Wang and Phillip Isola},
      year={2022},
      eprint={2005.10242},
      archivePrefix={arXiv},
      primaryClass={cs.LG},
      url={https://arxiv.org/abs/2005.10242}, 
}

@misc{ai4shipwrecks,
    title   = {Machine Learning for Shipwreck Segmentation from Side Scan Sonar Imagery: Dataset and Benchmark},
    author  = {Sethuraman, Advaith V. and Sheppard, Anja and Bagoren, Onur and Pinnow, Christopher and Anderson, Jamey and Havens, Timothy C. and Skinner, Katherine A.},
    journal = {arXiv preprint arXiv:2401.14546},
    year    = {2024},
    doi     = {10.48550/arXiv.2401.14546},
    url     = {https://arxiv.org/abs/2401.14546}
}

@misc{cobb2022sassed,
  author  = {Cobb, J. Tory},
  title   = {Synthetic Aperture Sonar Seabed Environment Dataset ({SASSED})},
  year    = {2022},
  version = {V4},
  publisher = {Mendeley Data},
  doi     = {10.17632/s5j5gzr2vc.4},
  url     = {https://data.mendeley.com/datasets/s5j5gzr2vc/4}
}

@incollection{sss_book,
title = {Chapter 1 - Introduction},
booktitle = {Acquisition and Processing of Marine Seismic Data},
publisher = {Elsevier},
pages = {1-35},
year = {2018},
isbn = {978-0-12-811490-2},
doi = {https://doi.org/10.1016/B978-0-12-811490-2.00001-3},
url = {https://www.sciencedirect.com/science/article/pii/B9780128114902000013},
author = {Derman Dondurur},
}

@BOOK{sss_handbook,
	author = {Blondel, P. and Murton, B.J.},
	title = {Handbook of seafloor sonar imagery},
	year = {1997},
	journal = {Handbook of seafloor sonar imagery},
	url = {https://www.scopus.com/inward/record.uri?eid=2-s2.0-0031422050&partnerID=40&md5=ecb1a03ce5ac7ff380a2b6d941cc0fe0},
	type = {Book},
	publication_stage = {Final},
    publisher = {Scopus},
	source = {Scopus},
	note = {Cited by: 158}
}

@misc{cnn_inductive,
      title={Theoretical Analysis of Inductive Biases in Deep Convolutional Networks}, 
      author={Zihao Wang and Lei Wu},
      year={2024},
      eprint={2305.08404},
      archivePrefix={arXiv},
      primaryClass={cs.LG},
      url={https://arxiv.org/abs/2305.08404}, 
}

@article{kmeans,
  title={Cluster analysis of multivariate data : efficiency versus interpretability of classifications},
  author={E. W. Forgy},
  journal={Biometrics},
  year={1965},
  volume={21},
  pages={768-769},
  url={https://api.semanticscholar.org/CorpusID:118110564}
}

@ARTICLE{knn,
  author={Cover, T. and Hart, P.},
  journal={IEEE Transactions on Information Theory}, 
  title={Nearest neighbor pattern classification}, 
  year={1967},
  volume={13},
  number={1},
  pages={21-27},
  doi={10.1109/TIT.1967.1053964}}

@ARTICLE{linear_probe,
  author={Yuan, Guo-Xun and Ho, Chia-Hua and Lin, Chih-Jen},
  journal={Proceedings of the IEEE}, 
  title={Recent Advances of Large-Scale Linear Classification}, 
  year={2012},
  volume={100},
  number={9},
  pages={2584-2603},
  doi={10.1109/JPROC.2012.2188013}}

@ARTICLE{yang2022sss,
    AUTHOR={Yang, Dianyu  and Cheng, Chensheng  and Wang, Can  and Pan, Guang  and Zhang, Feihu },
    TITLE={Side-Scan Sonar Image Segmentation Based on Multi-Channel CNN for AUV Navigation},
    JOURNAL={Frontiers in Neurorobotics},
    VOLUME={Volume 16 - 2022},
    YEAR={2022},
    URL={https://www.frontiersin.org/journals/neurorobotics/articles/10.3389/fnbot.2022.928206},
    DOI={10.3389/fnbot.2022.928206},
    ISSN={1662-5218},
}

@misc{chen2023vitadapter,
      title={Vision Transformer Adapter for Dense Predictions}, 
      author={Zhe Chen and Yuchen Duan and Wenhai Wang and Junjun He and Tong Lu and Jifeng Dai and Yu Qiao},
      year={2023},
      eprint={2205.08534},
      archivePrefix={arXiv},
      primaryClass={cs.CV},
      url={https://arxiv.org/abs/2205.08534}, 
}

@misc{chen2017deeplab,
      title={Rethinking Atrous Convolution for Semantic Image Segmentation}, 
      author={Liang-Chieh Chen and George Papandreou and Florian Schroff and Hartwig Adam},
      year={2017},
      eprint={1706.05587},
      archivePrefix={arXiv},
      primaryClass={cs.CV},
      url={https://arxiv.org/abs/1706.05587}, 
}

@misc{wang2020hrnet,
      title={Deep High-Resolution Representation Learning for Visual Recognition}, 
      author={Jingdong Wang and Ke Sun and Tianheng Cheng and Borui Jiang and Chaorui Deng and Yang Zhao and Dong Liu and Yadong Mu and Mingkui Tan and Xinggang Wang and Wenyu Liu and Bin Xiao},
      year={2020},
      eprint={1908.07919},
      archivePrefix={arXiv},
      primaryClass={cs.CV},
      url={https://arxiv.org/abs/1908.07919}, 
}

@misc{ronneberger2015unet,
      title={U-Net: Convolutional Networks for Biomedical Image Segmentation}, 
      author={Olaf Ronneberger and Philipp Fischer and Thomas Brox},
      year={2015},
      eprint={1505.04597},
      archivePrefix={arXiv},
      primaryClass={cs.CV},
      url={https://arxiv.org/abs/1505.04597}, 
}

@misc{kim2022inspyre,
      title={Revisiting Image Pyramid Structure for High Resolution Salient Object Detection}, 
      author={Taehun Kim and Kunhee Kim and Joonyeong Lee and Dongmin Cha and Jiho Lee and Daijin Kim},
      year={2022},
      eprint={2209.09475},
      archivePrefix={arXiv},
      primaryClass={cs.CV},
      url={https://arxiv.org/abs/2209.09475}, 
}

\clearpage % Forces the appendix to start on a new, clean page

\begin{figure*}[!ht]
    \centering
    \includegraphics[width=0.9\textwidth]{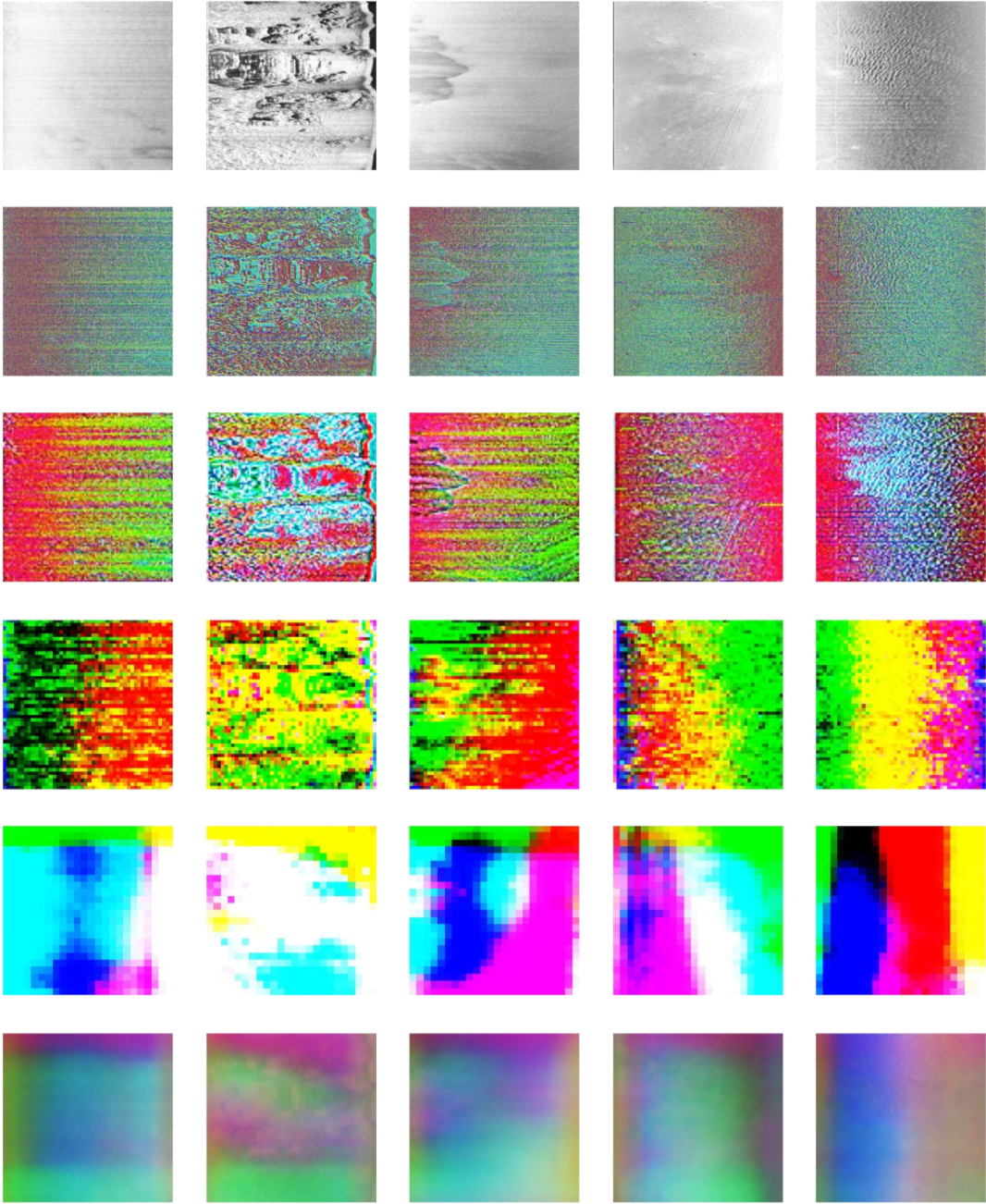}
    \caption{PCA visualization of feature maps across network stages. The first three principal components are mapped to RGB channels to visualize the high-dimensional feature spaces. Rows (from top to bottom) display: original image patches, intermediate outputs from Stages 1 through 4, and the final fused representation.}
    \label{fig:pca_vis}
\end{figure*}

\begin{figure*}[!ht]
    \centering
    \includegraphics[width=0.9\textwidth]{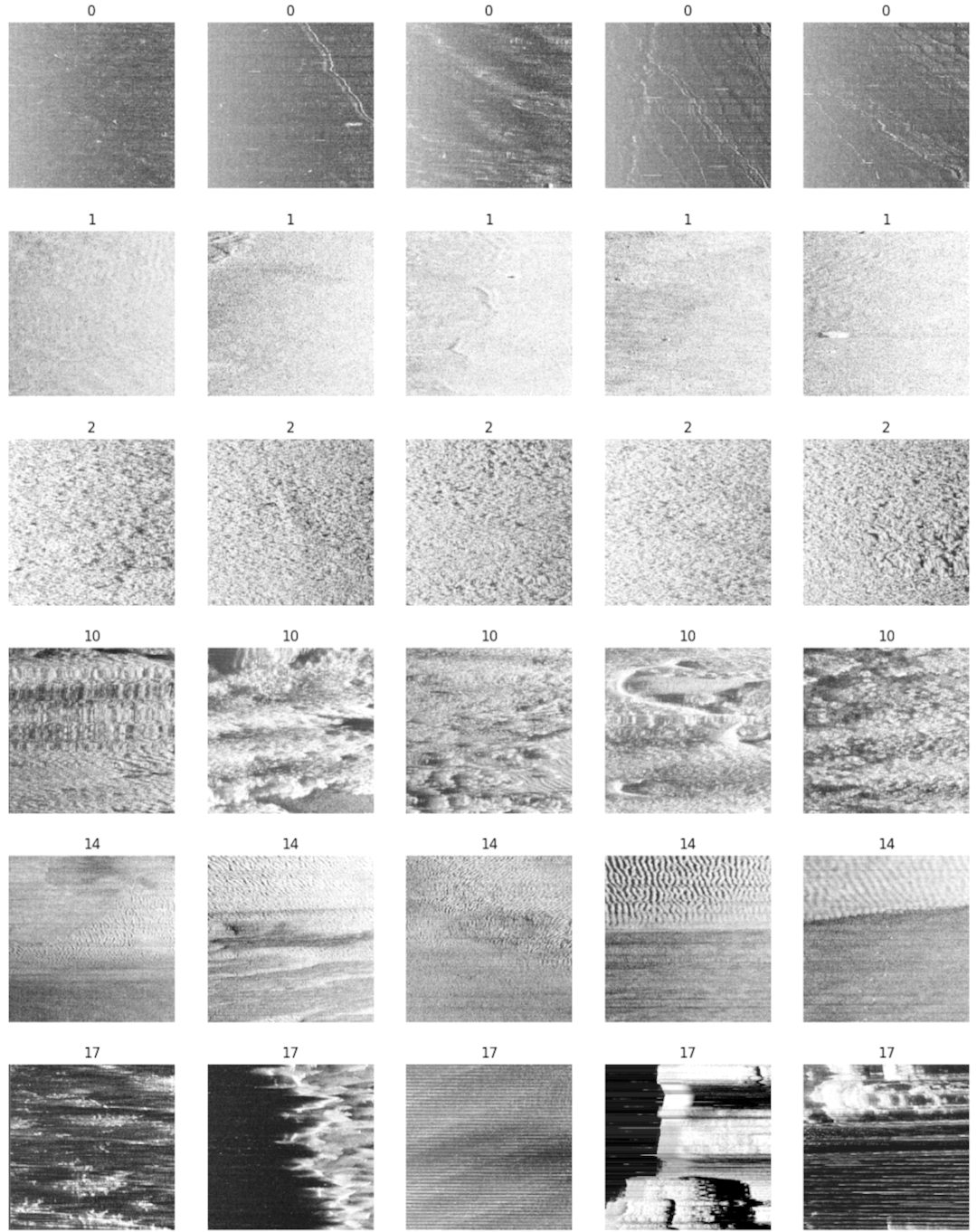}
    \caption{Top five matches for a sample of the centroids obtained by running K-Means clustering on 56 clusters. The centroid label is outlined on top of each respective image.}
    \label{fig:knn_vis1}
\end{figure*}

\begin{figure*}[!ht]
    \centering
    \includegraphics[width=0.9\textwidth]{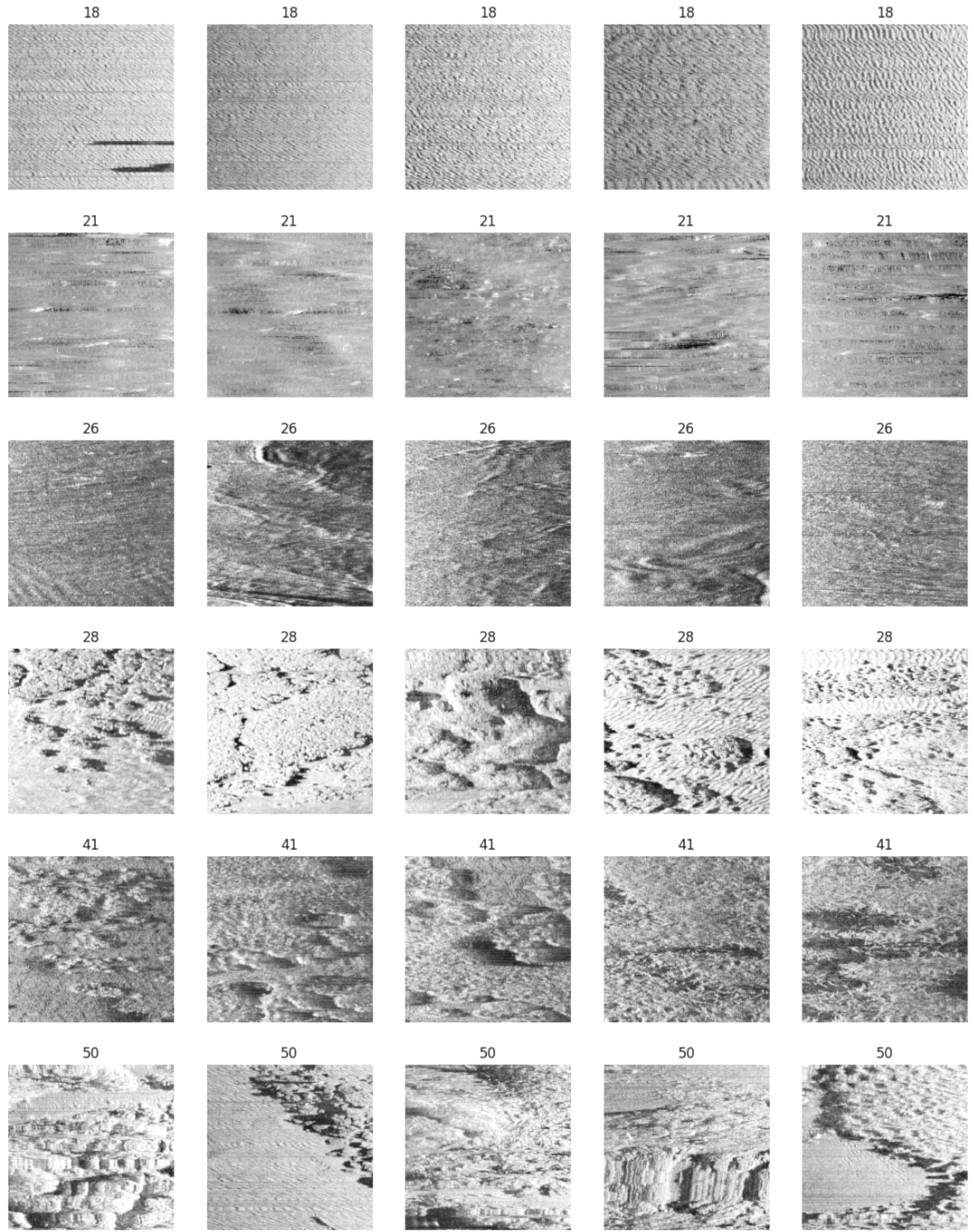}
    \caption{Continued: Top five matches for a sample of the centroids obtained using the K-Means clustering.}
    \label{fig:knn_vis2}
\end{figure*}

\end{document}